\documentclass[smallcondensed]{svjour3}     %

\usepackage{graphicx}
\usepackage{amsmath, amssymb, amsfonts}
\usepackage{slashbox}
\usepackage{float}
\usepackage[table, dvipsnames]{xcolor}
\usepackage{tikz}
\usepackage{tcolorbox}

\usepackage{pifont}%
\usepackage{bm}
\usepackage{booktabs}
\usepackage[sort&compress]{natbib}
\usepackage{textcomp}
\usepackage{booktabs}
\usepackage{adjustbox}
\usepackage{bbm}
\usepackage{colortbl}
\usepackage[normalem]{ulem}
\usepackage{multirow}
\usepackage{nicematrix}
\usepackage{array}
\usepackage{xparse}
\usepackage{xifthen}
\usepackage[final]{listings}
\usepackage[pagebackref=true,breaklinks=true]{hyperref}
\usepackage{ifdraft}
\usepackage[inline]{enumitem}
\usepackage{makecell}
\usepackage{mwe}
\usepackage[parfill]{parskip}
\usepackage{tasks}
\usepackage{mathtools}
\usepackage{mwe}
\usepackage[a4paper, total={6in, 8in}]{geometry}
\usepackage{marvosym}
\usepackage{scalerel}
\usepackage{academicons}
\usepackage{tablefootnote}
\usepackage{footmisc}

\newcommand*{\belowrulesepcolor}[1]{%
  \noalign{%
    \kern-\belowrulesep
    \begingroup
      \color{#1}%
      \hrule height\belowrulesep
    \endgroup
  }%
}

\newcommand*{\aboverulesepcolor}[1]{%
  \noalign{%
    \begingroup
      \color{#1}%
      \hrule height\aboverulesep
    \endgroup
    \kern-\aboverulesep
  }%
}

\hypersetup{final = true}
\hypersetup{
    colorlinks = true,
    linkcolor={red!70!black},
    citecolor={blue!70!black},
    urlcolor={blue!80!black}
}
\tcbuselibrary{most}
\usetikzlibrary{calc, decorations.pathreplacing, arrows, positioning}

\DeclareMathOperator*{\argmax}{arg\,max}
\DeclareMathOperator*{\argmin}{arg\,min}
\newcommand{\cmark}{\ding{51}}%
\newcommand{\xmark}{\ding{55}}%

\newcommand{\hypergeo}[1]{\text{Hypergeometric}(\ensuremath{#1})}
\newcommand{\EE}{\mathbf{E}}
\newcommand{\Var}{\mathbf{Var}}

\newcommand{\PP}{\mathbb{P}}
\newcommand{\MT}{\lfloor M\cdot\theta  \rceil}

\newcommand{\Hab}[3]{f_{X_#1}\left(#2, #3\right)}
\newcommand{\Zab}[3]{X_{#1}\left(#2, #3\right)}
\newcommand{\RZab}[3]{\mathcal{R}\left(\Zab{#1}{#2}{#3}\right)}

\newcommand{\THS}{\theta^*}
\newcommand{\THSSPACE}{\Theta^*}

\newcommand{\thmin}{\theta_{\text{min}}}
\newcommand{\thmax}{\theta_{\text{max}}}

\newcommand{\thsmin}{\THS_{\text{min}}}
\newcommand{\thsmax}{\THS_{\text{max}}}
\newcommand{\MTS}{M \cdot \THS}

\newcommand{\THMAX}[1]{\max_{\theta \in [0,1]} \left( #1 \right)}

\newcommand{\THARGMAX}[1]{\argmax_{\theta \in [0,1]} \left( #1 \right)}
\newcommand{\THSARGMAX}[1]{\argmax_{\THS \in \THSSPACE} \left\{ #1 \right\}}
\newcommand{\THMIN}[1]{\min_{\theta \in [0,1]} \left( #1 \right)}

\newcommand{\THARGMIN}[1]{\argmin_{\theta \in [0,1]} \left( #1 \right)}
\newcommand{\THSARGMIN}[1]{\argmin_{\THS \in \THSSPACE} \left\{ #1 \right\}}

\newcommand{\TP}{\text{TP}_{\theta}}
\newcommand{\TPS}{\text{TP}_{\THS}}
\newcommand{\TN}{\text{TN}_{\theta}}
\newcommand{\TNS}{\text{TN}_{\THS}}
\newcommand{\FP}{\text{FP}_{\theta}}
\newcommand{\FPS}{\text{FP}_{\THS}}
\newcommand{\FN}{\text{FN}_{\theta}}
\newcommand{\FNS}{\text{FN}_{\THS}}
\newcommand{\TPR}{\text{TPR}_{\theta}}
\newcommand{\TPRS}{\text{TPR}_{\THS}}
\newcommand{\TNR}{\text{TNR}_{\theta}}
\newcommand{\TNRS}{\text{TNR}_{\THS}}
\newcommand{\FPR}{\text{FPR}_{\theta}}
\newcommand{\FPRS}{\text{FNR}_{\THS}}
\newcommand{\FOR}{\text{FOR}_{\theta}}

\newcommand{\FNR}{\text{FNR}_{\theta}}
\newcommand{\FNRS}{\text{FNR}_{\THS}}
\newcommand{\TS}{\text{TS}_{\theta}}
\newcommand{\TSS}{\text{TS}_{\THS}}
\newcommand{\PT}{\text{PT}_{\theta}}
\newcommand{\GONE}{\text{FM}_{\theta}} %
\newcommand{\GONES}{\text{FM}_{\THS}}
\newcommand{\GTWO}{\text{G}^{(2)}_{\theta}}
\newcommand{\FB}{\text{F}^{(\beta)}_{\theta}}
\newcommand{\FBS}{\text{F}^{(\beta)}_{\THS}}
\newcommand{\PPV}{\text{PPV}_{\theta}}
\newcommand{\NPV}{\text{NPV}_{\theta}}
\newcommand{\J}{\text{J}_{\theta}}
\newcommand{\MK}{\text{MK}_{\theta}}
\newcommand{\ACC}{\text{Acc}_{\theta}}
\newcommand{\ACCS}{\text{Acc}_{\THS}}
\newcommand{\FDR}{\text{FDR}_{\theta}}
\newcommand{\BACC}{\text{BAcc}_{\theta}}

\newcommand{\MCC}{\text{MCC}_{\theta}}

\newcommand{\KT}{\kappa_{\theta}}
\newcommand{\PO}{P_o^{\theta}}
\newcommand{\PE}{P_e^{\theta}}

\newcommand{\TAC}[1]{\sigma_{#1}}

\newcommand{\MP}{M - P}
\newcommand{\MPP}{\left ( M - P \right )}

\newcommand{\FancyName}{DD}

\newcommand{\DDmax}{\Delta_{\max}}
\newcommand{\DDmin}{\Delta_{\min}}

\newcommand{\reColored}[2]{{\hypersetup{linkcolor=#1}\ref{#2}}}
\definecolor{lavendermagenta}{rgb}{0.93, 0.51, 0.93}

\usetikzlibrary{svg.path}

\definecolor{orcidlogocol}{HTML}{A6CE39}
\tikzset{
  orcidlogo/.pic={
    \fill[orcidlogocol] svg{M256,128c0,70.7-57.3,128-128,128C57.3,256,0,198.7,0,128C0,57.3,57.3,0,128,0C198.7,0,256,57.3,256,128z};
    \fill[white] svg{M86.3,186.2H70.9V79.1h15.4v48.4V186.2z}
                 svg{M108.9,79.1h41.6c39.6,0,57,28.3,57,53.6c0,27.5-21.5,53.6-56.8,53.6h-41.8V79.1z M124.3,172.4h24.5c34.9,0,42.9-26.5,42.9-39.7c0-21.5-13.7-39.7-43.7-39.7h-23.7V172.4z}
                 svg{M88.7,56.8c0,5.5-4.5,10.1-10.1,10.1c-5.6,0-10.1-4.6-10.1-10.1c0-5.6,4.5-10.1,10.1-10.1C84.2,46.7,88.7,51.3,88.7,56.8z};
  }
}

\newcommand\orcidicon[1]{\href{https://orcid.org/#1}{\mbox{\scalerel*{
    \begin{tikzpicture}[yscale=-1,transform shape]
    \pic{orcidlogo};
    \end{tikzpicture}
    }{|}}}
    }

\begin{document}

\title{The Dutch Draw: Constructing a Universal Baseline for Binary Prediction Models}

\titlerunning{The Dutch Draw}        %

\author{Etienne van de Bijl\textsuperscript{1, \Letter, \orcidicon{0000-0002-3245-4774}} \and
  Jan~Klein\textsuperscript{1, \Letter, \orcidicon{0000-0002-1777-194X}}
  \and
  Joris~Pries\textsuperscript{1, \Letter, \orcidicon{0000-0002-4429-6531}}
  \and
  Sandjai~Bhulai\textsuperscript{2, \orcidicon{0000-0003-1124-8821}} \and
  Mark~Hoogendoorn\textsuperscript{3,  \orcidicon{0000-0003-3356-3574}} \and
  Rob van der Mei\textsuperscript{1, \orcidicon{0000-0002-5685-5310}}
}

\authorrunning{E.\ v.d.\ Bijl, J.\ Klein, J.\ Pries et al.} %

\institute{ \at \textsuperscript{\Letter} Equal contribution and corresponding authors.
  \email{evdb@cwi.nl, j.g.klein@cwi.nl, joris.pries@cwi.nl}\\
  \at \textsuperscript{1}
  Centrum Wiskunde \& Informatica, Department of Stochastics\\
  Science Park 123\\
  1098 XG, Amsterdam, Netherlands\\
  \at \textsuperscript{2}
  Vrije Universiteit, Department of Mathematics \\
  De Boelelaan 1111 \\
  1081 HV, Amsterdam, Netherlands\\
  \at \textsuperscript{3}
  Vrije Universiteit, Department of Computer Science \\
  De Boelelaan 1111 \\
  1081 HV, Amsterdam, Netherlands\\
}

\maketitle

\begin{abstract}
  Novel prediction methods should always be compared to a baseline to know how well they perform. Without this frame of reference, the performance score of a model is basically meaningless. What does it mean when a model achieves an $F_1$ of 0.8 on a test set? A proper baseline is needed to evaluate the `goodness' of a performance score. Comparing with the latest state-of-the-art model is usually insightful. However, being state-of-the-art can change rapidly when newer models are developed. Contrary to an advanced model, a simple dummy classifier could be used. However, the latter could be beaten too easily, making the comparison less valuable. This paper presents a universal baseline method for all \emph{binary classification} models, named the \emph{Dutch Draw} (DD). This approach weighs simple classifiers and determines the best classifier to use as a baseline. We theoretically derive the DD baseline for many commonly used evaluation measures and show that in most situations it reduces to (almost) always predicting either zero or one. Summarizing, the DD baseline is:
  \begin{enumerate*}%
    \item [(1)] \emph{general}, as it is applicable to all binary classification problems;
    \item [(2)] \emph{simple}, as it is quickly determined without training or parameter-tuning;
    \item [(3)] \emph{informative}, as insightful conclusions can be drawn from the results.
  \end{enumerate*}
  The DD baseline serves two purposes. First, to enable comparisons across research papers by this robust and universal baseline. Secondly, to provide a sanity check during the development process of a prediction model. It is a major warning sign when a model is outperformed by the DD baseline.
\end{abstract}

\section{Introduction} \label{sec:introduction}

A typical data science project can be crudely simplified to the following steps: (1) comprehending the problem context, (2) understanding the data, (3) preparing the data, (4) modeling, (5) evaluating the model, and \textit{(6)} deploying the model \citep{Wirth2000}. Before deploying a new model, it should be tested whether it meets certain predefined success criteria. A baseline plays an essential role in this evaluation, as it gives an indication of the actual performance of a model.

However, which baseline should be selected? A good baseline is desirable, but what explicitly makes a baseline `good'? Comparing with the latest state-of-the-art model is usually insightful. However, being state-of-the-art can change rapidly when newer models are developed. Reproducibility of a model is also often a problem, because code is not published or large amounts of computational resources are required to retrain the model. These aspects make it hard or even impossible to compare older results with newer research. Nevertheless, it is important to stress that the comparison with a state-of-the-art model still has merit. However, we are pleading for an \emph{additional} universal baseline that can be computed quickly and can make it possible to compare results across research domains and papers. With that aim in mind, we outline three principal properties that any universal baseline should have: \emph{generality}, \emph{simplicity}, and \emph{informativeness}.

\paragraph{Generality}
In research, a new model is commonly compared to a limited number of existing models that are used in the same field. Although these are usually carefully selected, they are still subjectively chosen. Take binary classification, in which the objective is to label each observation either zero or one. Here, one could already select a decision tree \citep{Min2009}, random forest \citep{Couronne2018}, variants of naive Bayes \citep{Wang2012}, $k$-nearest neighbors \citep{Araujo2017}, support vector machine \citep{RaeisiShahraki2017}, neural network \citep{Sundarkumar2013}, or logistic regression model \citep{Sergioli2019} to evaluate the performance. These models are often trained specifically for a problem instance with parameters tuned for optimal performance in that specific case. Hence, these methods are not general. One could not take a decision tree that is used for determining bankruptcy \citep{Min2009} and use it as a baseline for a pathological voice detection problem \citep{Muhammad2014}. At least structural adaptations and retraining are necessary. A good standard baseline should be applicable to all binary classification problems, irrespective of the domain.

\paragraph{Simplicity}
An additional universal baseline should not be too complex. But, it is hard to determine for a measure if a baseline is too complex or not. Essentially, two components are critical in our view: \begin{enumerate*}[label=(\arabic*)]
  \item \emph{computational time} and
  \item \emph{explainability}.
\end{enumerate*}
It is necessary for practical applications that the baseline can be determined relatively fast. For example, training a neural network many times to generate an average baseline or optimizing the parameters of a certain model could take too much valuable time. Secondly, if a baseline is very complex, it can be harder to draw meaningful conclusions. Is it expected that a new model is outperformed by this ingeniously complicated baseline, or is it exactly what one would expect? This leads to the last property of a good standard baseline.

\paragraph{Informativeness}
Our baseline should be informative. When a method achieves a score higher or lower than the baseline, clear conclusions need to be drawn. Is it obvious that the baseline should be beaten? Consider the athletic event \emph{high jump}, where an athlete needs to jump over a bar at a specific height. If the bar is set too low, anyone can jump over it. If the bar is too high, no one makes it. Both situations do not give us any additional information to distinguish a professional athlete from a regular amateur. The bar should be placed at a height where the professional could obviously beat it, but the amateur can not. Drawing from this analogy, a baseline should be obviously beaten by any developed model. If not, this should be considered a major warning sign.

Our research focuses on finding such a general, simple and informative baseline for \emph{binary classification} problems. Although we focus on these type of problems, the three properties should also hold for constructing baselines in other supervised learning problems, such as multiclass classification and regression. Two methods that immediately come to mind are \emph{dummy classifiers} and \emph{optimal threshold classifiers}. They could be ideal candidates for our additional universal baseline.

\paragraph{Dummy classifier} A dummy classifier is a \emph{non-learning} model that makes predictions following a simple set of rules. For example, always predicting the \emph{most frequent} class label or predicting each class with some probability. A dummy classifier is simple and general, however it is not always informative. The information gained by performing better than a simple dummy classifier can even be zero. With the plethora of dummy classifiers, selection is also still arbitrary and questionable.

\paragraph{Optimal threshold classifier}
\cite{Koyejo2014} determined for a large family of binary performance measures that the optimal classifier consists of a sign function with a threshold tailored to each specific measure. To determine the optimal classifier, it is necessary to know or approximate $\PP(Y=1|X=x)$, which is the probability that the binary label $Y$ is 1 given the features $X=x$. \cite{Lipton2014} had a similar approach, but they only focus on the $F_1$ score. The conditional probabilities need to be learned from training data. However, this leads to arbitrary selections, as a model is necessary to approximate these probabilities. It is a clever approach, but unfortunately there is no clear-cut best approximation model for different research domains. If the approximation model is not accurate, the optimal classifier is based on wrong information, which makes it hard to draw meaningful conclusions from this approach.

Both the dummy classifier and the optimal threshold classifier have their strengths and weaknesses. In this paper, we introduce a novel baseline approach, called the \emph{Dutch Draw} (DD). The DD eliminates these weaknesses, whilst keeping their strengths. The DD can be seen as a dummy classifier on steroids. Instead of arbitrarily choosing a dummy classifier, we mathematically derive which classifier, from a family of classifiers, has the best expected performance. Also, this expected performance can be directly determined, making it very fast to obtain the baseline. The \FancyName{} baseline is:
\begin{enumerate*}%
  \item [(1)] applicable to any binary classification problem;
  \item [(2)] reproducible;
  \item [(3)] simple;
  \item [(4)] parameter-free;
  \item [(5)] more informative than any single dummy baseline;
  \item [(6)] and an explainable minimal requirement for any new model.
\end{enumerate*}
This makes the \FancyName{} an ideal candidate for a universal baseline in binary classification.

Our contributions are as follows:
\begin{enumerate*}[label=(\arabic*)]
  \item we introduce the \FancyName{} and explain why this method produces a universal baseline that is general, simple and informative for all binary classification problems;
  \item we provide the mathematical properties of the \FancyName{} for many evaluation measures and summarize them in several tables;
  \item we demonstrate how the DD baseline can be used in practice to identify which models should definitely be reconsidered;
  \item and we made the \FancyName{} available in a Python package.\footnote{\url{https://github.com/joris-pries/DutchDraw}}
\end{enumerate*}

\section{Preliminaries} \label{sec:preliminaries}

Before formulating the \FancyName{}, we need to introduce necessary notation, and simultaneously, provide elementary information on binary classification. This is required to explain how binary models are evaluated. Then, we discuss how performance measures are constructed for binary classification and we examine the ones that are most commonly used.

\subsection{Binary classification} \label{subsec:binaryclassification}

The goal of \emph{binary classification} is to learn (from a dataset) the relationship between the input variables and the binary output variable.
When the dataset consists of $M \in \mathbb{N}_{>0}$ observations, let $\mathcal{M} := \{1, \dots, M\}$ be the set of observation indices. Each instance, denoted by $\mathbf{x}_i$, has $K \in \mathbb{N}_{>0}$ explanatory feature values. These features can be categorical or numerical. Without loss of generality, we assume that $\mathbf{x}_i \in \mathbb{R}^K$ for all $i \in \mathcal{M}$. Moreover, each observation has a corresponding output value $y_i \in \{0,1\}$. Now, let $\mathbf{X} := [\mathbf{x}_1 \dots \mathbf{x}_M]^T \in \mathbb{R}^{M \times K}$ denote the matrix with all observations and their explanatory feature values and let $\mathbf{y} = (y_1, \dots, y_M)\in \{0,1\}^M$ be the response vector. The complete dataset is then represented by $(\mathbf{X}, \mathbf{y})$. We call the observations with response value $1$ `positive', while the observations with response value $0$ are `negative'. Let $P$ denote the number of positives and $N$ the number of negatives. Note that by definition $P + N = M$ must hold.

\subsection{Evaluation measures} \label{subsec:evaluation_measures}

An \textit{evaluation measure} quantifies the prediction performance of a trained model. We categorize the evaluation measures into two groups: \textit{base measures} and \textit{performance measures} \citep{Canbek2017}. Since there are two possible values for both the predicted and the true classes in binary classification, there are four base measures: the number of true positives $\text{(TP)}$, false positives $\text{(FP)}$, false negatives $\text{(FN)}$ and true negatives $\text{(TN)}$. Performance measures are a function of one or more these four base measures. To shorten notation, let $\hat{P} := \text{TP} + \text{FP}$ and $\hat{N} := \text{TN} + \text{FN}$ denote the number of positively and negatively predicted instances respectively.

All considered performance measures and base measures are shown in Table~\ref{Tab: Definition and Domain Measures}. Also their abbreviations, possibly alternative names, their definitions and corresponding codomains are presented in Table~\ref{Tab: Definition and Domain Measures}. The codomains show in what set the measure can theoretically take values (without considering the exact values of $P$, $N$, $\hat{P}$ and $\hat{N}$). In Sec.~\ref{sec:methodology}, the case-specific codomains are provided when we discuss the evaluation measures in more detail. Finally, note that the list is not exhaustive, but it contains most of the commonly used evaluation measures.

\begin{table}
  \begin{center}
    \bgroup
    \def\arraystretch{1.8}
    \caption{\textbf{\emph{Definitions and codomains of evaluation measures}}} %
    \label{Tab: Definition and Domain Measures}

    \begin{adjustbox}{width=\linewidth,keepaspectratio}
      \rowcolors{2}{white}{gray!25}
      \begin{tabular}{l>{$}l<{$}>{$}l<{$}} \toprule
        \textbf{Measure}                                                                        & \text{\bf Definition}                                                                                                                                    & \text{\bf Codomain} \\ \midrule
          \belowrulesepcolor{gray!25}
        True Positives (\hyperref[subsec: Appendix TP]{TP})                                     & \text{TP}                                                                                                                                                & \mathbb{N}_0        \\
        True Negatives (\hyperref[subsec: Appendix TN]{TN})                                     & \text{TN}                                                                                                                                                & \mathbb{N}_0        \\
        False Negatives (\hyperref[subsec: Appendix FN]{FN})                                    & \text{FN}                                                                                                                                                & \mathbb{N}_0        \\
        False Positives (\hyperref[subsec: Appendix FP]{FP})                                    & \text{FP}                                                                                                                                                & \mathbb{N}_0        \\
        True Positive Rate (\hyperref[subsec: Appendix TPR]{TPR}), Recall, Sensitivity          & \text{TPR} = \frac{\text{TP}}{\text{P}}                                                                                                                  & [ 0,1 ]             \\
        True Negative Rate (\hyperref[subsec: Appendix TNR]{TNR}), Specificity, Selectivity     & \text{TNR} = \frac{\text{TN}}{\text{N}}                                                                                                                  & [ 0,1 ]             \\
        False Negative Rate (\hyperref[subsec: Appendix FNR]{FNR}), Miss Rate                   & \text{FNR} = \frac{\text{FN}}{\text{P}}                                                                                                                  & [ 0,1 ]             \\
        False Positive Rate (\hyperref[subsec: Appendix FPR]{FPR}), Fall-out                    & \text{FPR} = \frac{\text{FP}}{\text{N}}                                                                                                                  & [ 0,1 ]             \\
        Positive Predictive Value (\hyperref[subsec: Appendix PPV]{PPV}), Precision             & \text{PPV} = \frac{\text{TP}}{\hat{P}}                                                                                                                   & [ 0,1 ]             \\
        Negative Predictive Value (\hyperref[subsec: Appendix NPV]{NPV})                        & \text{NPV} = \frac{\text{TN}}{\hat{N}}                                                                                                                   & [ 0,1 ]             \\
        False Discovery Rate (\hyperref[subsec: Appendix FDR]{FDR})                             & \text{FDR} = \frac{\text{FP}}{\hat{P}}                                                                                                                   & [ 0,1 ]             \\
        False Omission Rate (\hyperref[subsec: Appendix FOR]{FOR})                              & \text{FOR} = \frac{\text{FN}}{\hat{N}}                                                                                                                   & [ 0,1 ]             \\
        $F_{\beta}$ score (\hyperref[subsec: Appendix FBETA]{$F_{\beta}$})                      & F_{\beta} = (1+\beta^2)/\left(\frac{1}{\text{PPV}} + \frac{\beta^2}{\text{TPR}}\right)                                                                   & [ 0,1 ]             \\
        Youden's J Statistic/Index (\hyperref[subsec: Appendix J]{J}), (Bookmaker) Informedness & \text{J} =  \text{TPR} + \text{TNR} - 1                                                                                                                  & [-1,1 ]             \\
        Markedness (\hyperref[subsec: Appendix MK]{MK})                                         & \text{MK} = \text{PPV} + \text{NPV} - 1                                                                                                                  & [-1,1 ]             \\
        Accuracy (\hyperref[subsec: Appendix ACC]{Acc})                                         & \text{Acc} = \frac{\text{TP} + \text{TN}}{M}                                                                                                             & [ 0,1 ]             \\
        Balanced Accuracy (\hyperref[subsec: Appendix BACC]{BAcc})                              & \text{BAcc} = \frac{1}{2}(\text{TPR} + \text{TNR})                                                                                                       & [ 0,1 ]             \\
        Matthews Correlation Coefficient (\hyperref[subsec: Appendix MCC]{MCC})                 & \text{MCC} = \frac{\text{TP} \cdot \text{TN} - \text{FP} \cdot \text{FN}}{\sqrt{\hat{P} \cdot \hat{N} \cdot \text{P} \cdot \text{N}}}                    & [-1,1 ]             \\
        Cohen's kappa (\hyperref[subsec: Appendix KAPPA]{$\kappa$})                             & \kappa = \frac{\text{P}_o - \text{P}_e}{1-\text{P}_e}, \text{ with } \text{P}_o = \text{Acc}, \text{P}_e = \frac{\hat{P} \cdot P + \hat{N} \cdot N}{M^2} & [-1,1 ]             \\
        Fowlkes-Mallows Index (\hyperref[subsec: Appendix FM]{FM}), G-mean 1                    & \text{FM} = \sqrt{\text{TPR} \cdot \text{PPV}}                                                                                                           & [ 0,1 ]             \\
        G-mean 2 (\hyperref[subsec: Appendix G2]{$\text{G}^{(2)}$})                             & \text{G}^{(2)} = \sqrt{\text{TPR} \cdot \text{TNR}}                                                                                                      & [ 0,1 ]             \\
        Prevalence Threshold (\hyperref[subsec: Appendix PT]{PT})                               & \text{PT} =\frac{\sqrt{\text{TPR} \cdot \text{FPR}} - \text{FPR}}{\text{TPR} - \text{FPR}}                                                               & [ 0,1 ]             \\
        Threat Score (\hyperref[subsec: Appendix TS]{TS}), Critical Success Index               & \text{TS} =\frac{\text{TP}}{P + \text{FP}}                                                                                                               & [ 0,1 ]             \\  \aboverulesepcolor{gray!25} \bottomrule
      \end{tabular}
    \end{adjustbox}

    \egroup
  \end{center}
\end{table}

\subsubsection*{Ill-defined measures} \label{subsec:illdefinedmeasures}

Not every evaluation measure is well-defined. Often, the problem occurs due to division by zero. For example, the \emph{True Positive Rate} (TPR) defined as $\text{TPR} = \text{TP}/P$ cannot be calculated whenever $P=0$. Therefore, we have made assumptions for the allowed values of $P$, $N$, $\hat{P}$ and $\hat{N}$. These are shown in Table~\ref{tab: Assumptions P and N}. One exception is the \emph{Prevalence Threshold} (PT) \citep{Balayla2020}, where the denominator is zero if TPR is equal to the False Positive Rate (defined as $\text{FPR} = \text{FP}/N$). Depending on the classifier, this situation could occur regularly. Therefore, PT is omitted throughout the rest of this research.

\renewcommand{\xmark}{-}
\renewcommand{\cmark}{$>0$}

\begin{table}
  \centering
  \def\arraystretch{1.5}
  \caption{\textbf{\emph{Assumptions on domains $P$, $N$, $\hat{P}$ and $\hat{N}$:}} Some measures are not defined if $P$, $N$, $\hat{P}$ or $\hat{N}$ is equal to zero. These domain requirements are therefore necessary (always $M > 0$).}
  \label{tab: Assumptions P and N}
  \rowcolors{2}{white}{gray!25}
  \begin{tabular}{lllll} \toprule
                                                                                                                               & \multicolumn{4}{c}{\bf Domain requirement for:}                                  \\ \cmidrule(lr){2-5}
    \rowcolor{white} \bf Measure                                                                                               & $P$                                             & $N$    & $\hat{P}$ & $\hat{N}$ \\ \midrule
    \belowrulesepcolor{gray!25}
    \hyperref[subsec: Appendix TP]{TP}, \hyperref[subsec: Appendix TN]{TN}, \hyperref[subsec: Appendix FN]{FN}, \hyperref[subsec: Appendix FP]{FP},
    \hyperref[subsec: Appendix ACC]{Acc}, \hyperref[subsec: Appendix KAPPA]{$\kappa$}                                          & \xmark                                          & \xmark & \xmark    & \xmark    \\
    \hyperref[subsec: Appendix TPR]{TPR}, \hyperref[subsec: Appendix FNR]{FNR}, \hyperref[subsec: Appendix TS]{TS}             & \cmark                                          & \xmark & \xmark    & \xmark    \\
    \hyperref[subsec: Appendix TNR]{TNR}, \hyperref[subsec: Appendix FPR]{FPR}                                                 & \xmark                                          & \cmark & \xmark    & \xmark    \\
    \hyperref[subsec: Appendix PPV]{PPV}, \hyperref[subsec: Appendix FDR]{FDR}                                                 & \xmark                                          & \xmark & \cmark    & \xmark    \\
    \hyperref[subsec: Appendix NPV]{NPV}, \hyperref[subsec: Appendix FOR]{FOR}                                                 & \xmark                                          & \xmark & \xmark    & \cmark    \\
    \hyperref[subsec: Appendix FBETA]{$F_{\beta}$}, \hyperref[subsec: Appendix FM]{FM}                                         & \cmark                                          & \xmark & \cmark    & \xmark    \\
    \hyperref[subsec: Appendix J]{J}, \hyperref[subsec: Appendix BACC]{BAcc}, \hyperref[subsec: Appendix G2]{$\text{G}^{(2)}$} & \cmark                                          & \cmark & \xmark    & \xmark    \\
    \hyperref[subsec: Appendix MK]{MK}                                                                                         & \xmark                                          & \xmark & \cmark    & \cmark    \\
    \hyperref[subsec: Appendix MCC]{MCC}                                                                                       & \cmark                                          & \cmark & \cmark    & \cmark    \\ \aboverulesepcolor{gray!25} \bottomrule
  \end{tabular}
\end{table}

\section{Dutch Draw (DD)} \label{sec:methodology}

In this section, we introduce the \FancyName{} framework and discuss how this method is able to provide a universal baseline for any evaluation measure. This baseline is general, simple, and informative, which is crucial for a good baseline, as we explained in Sec.~\ref{sec:introduction}. First, we provide the family of \FancyName{} classifiers, and thereafter explain how the optimal classifier generates the baseline.

\subsection{Dutch Draw classifiers} \label{subsec:shuffle}

The goal of our research is to provide a universal baseline for any evaluation measure in binary classification. The \FancyName{} baseline comes from choosing the best \FancyName{} classifier. Before we discuss what `best' actually entails, we have to define the \FancyName{} classifier in general. This is the function $\TAC{\theta}: \mathbb{R}^{M \times K} \rightarrow \{0,1\}^M$ with input an evaluation dataset with $M$ observations and $K$ feature values per observation. The function generates the predictions for these observations by outputting a vector of $M$ binary predictions. It is described in words as:
\begin{align*}
  \TAC{\theta}(\mathbf{X}) := & \{\text{take a random sample without replacement of size $\MT$}                \\
                              & \text{ of rows from $\mathbf{X}$ and assign $1$ to these observations and $0$} \\
                              & \text{ to the remaining rows}\}.
\end{align*}

Here, $\lfloor \cdot \rceil$ is the function that rounds its argument to the nearest integer. %
The parameter $\theta \in [0,1]$ controls what percentage of observations are predicted as positive.
The mathematical definition of $\TAC{\theta}$ is given by:
\begin{align*}
  \TAC{\theta}(\mathbf{X}) & := \left(\mathbf{1}_{E}(i)\right)_{i \in \mathcal{M}} \text{ with } E\subseteq \mathcal{M} \text{ uniformly drawn s.t. }|E|  = \MT,
\end{align*}
with $\left(\mathbf{1}_{E}(i)\right)_{i \in \mathcal{M}}$ the vector with ones in the positions in $E$ and zeroes elsewhere. Note that a classifier $\TAC{\theta}$ does not learn from the features in the data, just as a dummy classifier. The set of all \FancyName{} classifiers $\{\TAC{\theta}: \theta \in [0,1]\}$ is the complete family of models that classify a random sample of any size as positive.

Given a \FancyName{} classifier, the number of predicted positives $\hat{P}$ depends on $\theta$ and is given by $\hat{P}_{\theta} := \MT$ and the number of predicted negatives is $\hat{N}_{\theta} := M - \MT$. To be specific, these two numbers are integers, and thus, different values of $\theta$ can lead to the same value of $\hat{P}_{\theta}$. Therefore, we introduce the parameter $\THS := \frac{\MT}{M}$ as the discretized version of $\theta$. Furthermore, we define:
\begin{align*}
  \THSSPACE := \left\{\frac{\MT}{M}: \theta \in [0,1]\right\} = \left\{0, \frac{1}{M}, \dots, \frac{M-1}{M}, 1\right\}
\end{align*}
as the set of all unique values that $\THS$ can obtain for all $\theta \in [0,1]$.

Next, we derive mathematical properties of the \FancyName{} classifier for every evaluation measure in Table~\ref{Tab: Definition and Domain Measures} (except PT). Note that the \FancyName{} is stochastic, thus we examine the \emph{distribution} of the evaluation measure. Furthermore, we also determine the \emph{range} and \emph{expectation} of a \FancyName{} classifier.

\subsubsection{Distribution}
The distributions of the base measures (see Sec.~\ref{subsec:evaluation_measures}) are directly determined by $\TAC{\theta}$. Consider for example $\text{TP}$: the number of positive observations that are also predicted to be positive. In a dataset of $M$ observations with $P$ labeled positive, $\MT$ random observations are predicted as positive in the \FancyName{} approach. This implies that $\TP$ is hypergeometrically distributed with parameters $M$, $P$ and $\MT$, as the classifier randomly draws $\MT$ samples without replacement from a population of size $M$, where $P$ samples are labeled positive. Thus:
\begin{align*}
  \PP(\TP = s) = \begin{cases}\frac{\binom{P}{s} \cdot \binom{M - P}{\MT - s}}{\binom{M}{\MT}} & \text{if $s \in \mathcal{D}(\TP)$,} \\ 0 & \text{else,} \end{cases}
\end{align*}
where $\mathcal{D}(\TP)$ is the domain of $\TP$. The definition of this domain is given in Eq.~\eqref{eq: Domain TP}.

The other three base measures are also hypergeometrically distributed following similar reasoning. This leads to:
\begin{align*}
  \TP & \sim \hypergeo{M, P, \MT},   \\
  \FP & \sim \hypergeo{M, N, \MT},   \\
  \FN & \sim \hypergeo{M, P, M-\MT}, \\
  \TN & \sim \hypergeo{M, N, M-\MT}.
\end{align*}
Note that these random variables are not independent. In fact, they can all be written in terms of $\TP$. This is a crucial effect of the DD approach, as it reduces the formulations to only a function of a single variable. Consequently, most evaluation measures can be written as a linear combination of only $\TP$. With only one random variable, theoretical derivations and optimal classifiers can be determined. As mentioned before, $\TP + \FN = P$ and $\TN + \FP = N = M-P$, and we also have $\TP + \FP = \MT$, because this denotes the total number of positively predicted observations. These three identities are linear in $\TP$, thus each base measure can be written in the form $\Zab{\theta}{a}{b} := a\cdot\TP + b$ with $a, b \in \mathbb{R}$. Additionally, let $\Hab{\theta}{a}{b}$ be the probability distribution of $\Zab{\theta}{a}{b}$. Then, by combining the identities, we get:
\begin{align}
  \TP & = \TP                        %
  ,  \label{eq: TP in TP}\tag{B1}    \\
  \FP & = \hat{P}_{\theta} - \TP     %
  , \label{eq: FP in TP}\tag{B2}     \\
  \FN & = P - \TP                    %
  ,  \label{eq: FN in TP}\tag{B3}    \\
  \TN & = N - \hat{P}_{\theta} + \TP %
  ,  \label{eq: TN in TP}\tag{B4}
\end{align}
with $\hat{P}_{\theta} := \MT$. %

\paragraph{Example: distribution $F_{\beta}$ score}
To illustrate how the probability function $\Hab{\theta}{a}{b}$ can directly be derived, %
we consider the \emph{$F_{\beta}$ score} $\FB$ \citep{Chinchor1992}. It is the \emph{weighted harmonic average} between the \emph{True Positive Rate} ($\TPR$) and the \emph{Positive Predictive Value} ($\PPV$). The latter two performance measures are discussed extensively in \ref{subsec: Appendix TPR} and~\ref{subsec: Appendix PPV}, respectively. The $F_{\beta}$ score balances predicting the actual positive observations correctly ($\TPR$) and being cautious in predicting observations as positive ($\PPV$). The factor $\beta > 0$ indicates how much more $\TPR$ is weighted compared to $\PPV$. The $F_{\beta}$ score is commonly defined as:
\begin{equation*}
  \FB = \frac{1+\beta^2}{\frac{1}{\PPV} + \frac{\beta^2}{\TPR}}.
\end{equation*}
By substituting $\PPV$ and $\TPR$ by their definitions (see Table~\ref{Tab: Definition and Domain Measures}) and using Eq.~\eqref{eq: TP in TP} and~\eqref{eq: FP in TP}, we get:
\begin{align*}
  \FB = \frac{(1+\beta^2)\TP}{\beta^2 \cdot P + \MT}.
\end{align*}

Since $\PPV$ is only defined when $\hat{P}_{\theta} = \MT > 0$ and $\TPR$ is only defined when $P > 0$, we need for $\FB$ that both these restrictions hold. The definition of $\FB$ is linear in $\TP$ and can therefore be formulated as:
\begin{align*}
  \FB & = \Zab{\theta}{\frac{1+\beta^2}{\beta^2\cdot P + \MT}}{0}.
\end{align*}

\subsubsection{Range} \label{subsubsec: codomains evaluation measures}
The values that $\Zab{\theta}{a}{b}$ can attain depend on $a$ and $b$, and of course, on the domain of $\TP$. Without restriction, the maximum number that $\TP$ can be is $P$. Then, all positive observations are also predicted to be positive. However, when $\theta$ is small enough such that $\MT < P$, then only $\MT$ observations are predicted as positive. Consequently, $\TP$ can only reach the value $\MT$ in this case. Hence, in general, the upper bound of the domain of $\TP$ is $\min\{P, \MT\}$. The same reasoning holds for the lower bound: when $\theta$ is small enough, the minimum number of $\TP$ is 0, since all positive observations can be predicted as negative. However, when $\theta$ gets large enough, positive observations have to be predicted positive even if all $\MP$ negative observations are predicted positive. Thus, in general, the lower bound of the domain is $\max\{0, \MT - (M-P)\}$. Now, let $\mathcal{D}(\TP)$ be the domain of $\TP$, then:
\begin{align}
  \mathcal{D}(\TP) := \left\{i \in \mathbb{N}_0: \max\{0, \MT - (M-P)\} \leq i \leq \min\{P, \MT\}\right\}. \label{eq: Domain TP}
\end{align}
Consequently, the range of $\Zab{\theta}{a}{b}$ is given by
\begin{align}
  \RZab{\theta}{a}{b} := \left \{a \cdot i + b \right \}_{i \in \mathcal{D}(\TP)}. \label{eq: Range Zab}\tag{R}
\end{align}

\subsubsection{Expectation}
The introduction of $\Zab{\theta}{a}{b}$ allows us to write its expected value %
in terms of $a$ and $b$. This statistic is required to calculate the actual baseline. %
Since $\TP$ has a $\hypergeo{M, P, \MT}$ distribution, its expected value %
is known and given by
\begin{equation*}
  \EE[\TP] = \frac{\MT}{M} \cdot P. %
\end{equation*}
Next, we obtain the following general definition for the expectation %
of $\Zab{\theta}{a}{b}$:
\begin{align}
  \EE[\Zab{\theta}{a}{b}] & = a \cdot \EE[\TP] + b = a \cdot \frac{\MT}{M} \cdot P + b, \label{eq: Expectation Rule 1}\tag{\ensuremath{\square}} %
\end{align}
This rule is consistently used to determine the expectation for each measure.

\paragraph{Example: expectation $F_{\beta}$ score}
To demonstrate how the expectation is calculated for a performance measure, we again consider $\FB$.
It is linear in $\TP$ with $a = (1+\beta^2)/(\beta^2 \cdot P + \MT)$ and $b = 0$, and so, its expectation is given by:
\begin{align}
  \EE[\FB] & = \EE\left[\Zab{\theta}{\frac{1+\beta^2}{\beta^2\cdot P + \MT}}{0}\right] \stackrel{\eqref{eq: Expectation Rule 1}}{=} \frac{1+\beta^2}{\beta^2\cdot P + \MT} \cdot \EE[\TP] + 0 \nonumber \\
           & = \frac{\MT \cdot P \cdot (1+\beta^2)}{M\cdot(\beta^2 \cdot P + \MT)} \nonumber                                                                                                            \\
           & = \frac{(1+\beta^2)\cdot P\cdot \THS}{\beta^2\cdot P + M\cdot\THS}. \label{eq:fbetae}
\end{align}

A full overview of the distribution and mean of all considered base and performance measures is given in Table~\ref{tab: Properties baseline measures}. All the calculations performed to derive the corresponding distributions and expectations are provided in \hyperref[app: Appendix]{Appendix A}.

\begin{table}
  \centering
  \def\arraystretch{2.0}
  \caption{\textbf{\emph{Properties of performance measures for a \FancyName{} classifier:}}
    Expectation and distribution of each performance measure for a \FancyName{} classifier $\TAC{\theta}$ with $\THS =  \frac{\MT}{M}$.}
  \label{tab: Properties baseline measures}
  \rowcolors{2}{white}{gray!25}
  \begin{tabular}{llll} \toprule
                                                                                  &                                                               & \multicolumn{2}{l}{\bf Distribution $\bm{\Hab{\theta}{a}{b}}$}                                                        \\ \cmidrule{3-4}
    \rowcolor{white}  \multirow{-2}{*}{\bf Measure}                               & \multirow{-2}{*}{\bf Expectation}                             & $a$                                                            & $b$                                                  \\ \midrule
    \belowrulesepcolor{gray!25}
    \rule[-2.5ex]{0pt}{0pt}      \hyperref[subsec: Appendix TP]{TP}               & $ \THS \cdot P$                                               & $ 1  $                                                         & $ 0 $                                                \\
    \rule[-2.5ex]{0pt}{0pt}    \hyperref[subsec: Appendix TN]{TN}                 & $ (1-\THS) \MPP $                                             & $ 1  $                                                         & $ \MP - M\cdot\THS $                                 \\
    \rule[-2.5ex]{0pt}{0pt}    \hyperref[subsec: Appendix FN]{FN}                 & $ (1-\THS) P$                                                 & $ -1  $                                                        & $ P $                                                \\
    \rule[-2.5ex]{0pt}{0pt}        \hyperref[subsec: Appendix FP]{FP}             & $\THS \MPP $                                                  & $ -1  $                                                        & $ M\cdot\THS $                                       \\
    \rule[-2.5ex]{0pt}{0pt}        \hyperref[subsec: Appendix TPR]{TPR}           & $\THS $                                                       & $\frac{1}{P}$                                                  & $ 0 $                                                \\
    \rule[-2.5ex]{0pt}{0pt}        \hyperref[subsec: Appendix TNR]{TNR}           & $1-\THS $                                                     & $\frac{1}{\MP} $                                               & $ 1-\frac{M\cdot\THS}{\MP}$                          \\
    \rule[-2.5ex]{0pt}{0pt}    \hyperref[subsec: Appendix FNR]{FNR}               & $1 - \THS$                                                    & $-\frac{1}{P}$                                                 & $ 1 $                                                \\
    \rule[-2.5ex]{0pt}{0pt}        \hyperref[subsec: Appendix FPR]{FPR}           & $\THS $                                                       & $-\frac{1}{\MP}$                                               & $\frac{M\cdot\THS}{\MP}$                             \\
    \rule[-2.5ex]{0pt}{0pt}    \hyperref[subsec: Appendix PPV]{PPV}               & $\frac{P}{M} $                                                & $\frac{1}{M\cdot\THS}$                                         & $ 0 $                                                \\
    \rule[-2.5ex]{0pt}{0pt}        \hyperref[subsec: Appendix NPV]{NPV}           & $1 - \frac{P}{M}$                                             & $\frac{1}{M(1-\THS)}$                                          & $ 1 - \frac{P}{M(1-\THS)} $                          \\
    \rule[-2.5ex]{0pt}{0pt}        \hyperref[subsec: Appendix FDR]{FDR}           & $1 - \frac{P}{M}$                                             & $-\frac{1}{M\cdot\THS}$                                        & $1$                                                  \\
    \rule[-2.5ex]{0pt}{0pt}        \hyperref[subsec: Appendix FOR]{FOR}           & $\frac{P}{M} $                                                & $-\frac{1}{M(1-\THS)}$                                         & $\frac{P}{M(1-\THS)} $                               \\
    \rule[-2.5ex]{0pt}{0pt}        \hyperref[subsec: Appendix FBETA]{$F_{\beta}$} & $\frac{(1+ \beta ^2) \THS \cdot P}{\beta ^2 \cdot P + \MTS} $ & $\frac{1+\beta^2}{\beta^2 \cdot P + \MTS}$                     & $0$                                                  \\
    \rule[-2.5ex]{0pt}{0pt}    \hyperref[subsec: Appendix J]{J}                   & $0$                                                           & $\frac{M}{P\MPP}$                                              & $-\frac{M\cdot\THS}{\MP} $                           \\
    \rule[-2.5ex]{0pt}{0pt}    \hyperref[subsec: Appendix MK]{MK}                 & $0$                                                           & $\frac{1}{M\cdot\THS(1-\THS)}$                                 & $-\frac{P}{M(1-\THS)} $                              \\
    \rule[-2.5ex]{0pt}{0pt}    \hyperref[subsec: Appendix ACC]{Acc}               & $\frac{(1-\THS) \MPP + \THS \cdot P}{M} $                     & $\frac{2}{M}$                                                  & $1 - \THS - \frac{P}{M}$                             \\
    \rule[-2.5ex]{0pt}{0pt}\hyperref[subsec: Appendix BACC]{BAcc}                 & $\frac{1}{2} $                                                & $\frac{M}{2P\MPP}$                                             & $\frac{1}{2} - \frac{M\cdot\THS}{2\MPP}$             \\
    \rule[-2.5ex]{0pt}{0pt} \hyperref[subsec: Appendix MCC]{MCC}                  & $0$                                                           & $\frac{1}{\sqrt{P \MPP \THS(1-\THS)}}$                         & $-\frac{\sqrt{P \cdot \THS}}{\sqrt{\MPP (1-\THS)} }$ \\
    \rule[-2.5ex]{0pt}{0pt} \hyperref[subsec: Appendix KAPPA]{$\kappa$}           & $0$                                                           & $\frac{2}{P(1-\THS)+\MPP\THS}$                                 & ${-\frac{2\THS \cdot P}{P(1-\THS) + \MPP\THS}}$      \\
    \rule[-2.5ex]{0pt}{0pt} \hyperref[subsec: Appendix FM]{FM}                    & $\sqrt{\frac{\THS \cdot P}{M}}$                               & $ \frac{1}{\sqrt{P \cdot M \cdot \THS}}$                       & $ {0}$                                               \\
    \rule[-2.5ex]{0pt}{0pt} \hyperref[subsec: Appendix G2]{$\text{G}^{(2)}$}      & \xmark                                                        & Nonlinear in $\TP$                                             & Nonlinear in $\TP$                                   \\
    \rule[-2.5ex]{0pt}{0pt}    \hyperref[subsec: Appendix TS]{TS}                 & \xmark                                                        & Nonlinear in $\TP$                                             & Nonlinear in $\TP$                                   \\ \bottomrule
  \end{tabular}
  \vspace*{1 mm}

\end{table}

\subsection{Optimal Dutch Draw classifier} \label{sec: optimal dutch draw classifier}
Next, we discuss how the \FancyName{} baseline will ultimately be derived. In order to do so, an overview is presented in Fig.~\ref{fig:workflowdd}. Starting with the definition of the \FancyName{} classifiers in Sec.~\ref{subsec:shuffle} and determining their expectations for commonly used measures (see Table~\ref{tab: Properties baseline measures}), we are now able to identify the \emph{optimal} \FancyName{} classifier. Given a performance measure and dataset,
the optimal \FancyName{} classifier is found by optimizing (taking the minimum or maximum of) the associated expectation for $\theta \in [0,1]$.

\definecolor{persianpink}{rgb}{0.97, 0.5, 0.75}

\definecolor{color1}{rgb}{0.31, 0.78, 0.47}

\newtcolorbox{mybox}[2][]{
  box align = center,
  halign = center,
  width = 5.5 cm,
  colback=red!5!white,
  colframe=red!75!black,
  colbacktitle=red!85!black,enhanced,
  attach boxed title to top center={yshift=-2mm},
  title=#2,#1}

\begin{figure}[H]
  \begin{center}
    \resizebox*{0.7\linewidth}{!}{
      \begin{tikzpicture}

        \node[anchor = center] (first) at (0,0) {\begin{mybox}[colback=gray!15, colframe = color1, colbacktitle = color1]{\S\reColored{white}{subsec:shuffle}}
            All \FancyName{} classifiers
          \end{mybox}};

        \node[anchor = center] (second) at (7,0) {\begin{mybox}[colback=gray!15, colframe = color1, colbacktitle = color1]{\color{white} Table~\reColored{white}{tab: Properties baseline measures}}
            Expectations of all \FancyName{} classifiers
          \end{mybox}};

        \node[anchor = center] (third) at (0,-3) {\begin{mybox}[colback=gray!15, colframe = color1, colbacktitle = color1]{\S\reColored{white}{sec: optimal dutch draw classifier}}
            Optimal \FancyName{} classifier
          \end{mybox}};

        \node[anchor = center] (fourth) at (7, -3) {\begin{mybox}[colback=gray!15, colframe = color1, colbacktitle = color1]{\color{white} Table~\reColored{white}{tab: Optimal baseline measures}}
            \FancyName{} baseline
          \end{mybox}};

        \begin{scope}[transform canvas={yshift=-.7em}]

          \draw[->, ultra thick, shorten >=3pt, shorten <=3pt, persianpink] (first) -- node [above, midway]{(1)} (second);

          \draw[->, ultra thick, shorten >=3pt, shorten <=3pt, persianpink] (second) -- node [above, midway]{(2)} (third);

          \draw[->, ultra thick, shorten >=3pt, shorten <=3pt, persianpink] (third) -- node [above, midway]{(3)} (fourth);

        \end{scope}

      \end{tikzpicture}
    }

    \caption{\textbf{\emph{Road to \FancyName{} baseline:}} This is an overview of how the \FancyName{} baseline is determined. (1) all expectations are derived; (2) the expectation is maximized/minimized; (3) the performance of the best \FancyName{} classifier is the \FancyName{} baseline}
    \label{fig:workflowdd}

  \end{center}
\end{figure}
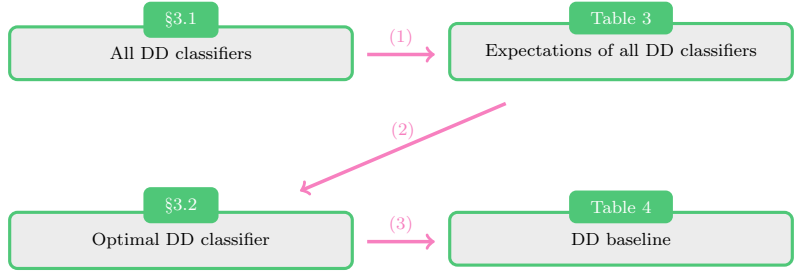

\subsubsection{Dutch Draw baseline} \label{subsec:optknowpm}
The optimal \FancyName{} classifiers and the corresponding \FancyName{} baseline can be found in Table~\ref{tab: Optimal baseline measures}. For many performance measures, it is optimal to always predict positive or negative. In some cases, this is not allowed due to ill-defined measures. Then, it is often optimal to only predict one sample differently. For several other measures, almost all parameter values give the optimal baseline. Next, we give an example to illustrate how the results of Table~\ref{tab: Optimal baseline measures} are derived.

\paragraph{Example: \FancyName{} baseline for the $F_{\beta}$ score} To determine the \FancyName{} baseline, the extreme values of the expectation $\EE[\FB]$ need to be identified. To do this, examine the following function $f: [0,1] \rightarrow [0,1]$ defined as:
\begin{equation*}
  f(t) = \frac{(1+\beta^2) \cdot P \cdot t}{\beta^2 \cdot P + M \cdot t}.
\end{equation*}
The relationship between $f$ and $\EE[\FB]$ is given as $f(\MT / M) = \EE[\FB]$. To find the extreme values, we have to look at the derivative of $f$:
\begin{equation*}
  \frac{\mathrm{d}f(t)}{\mathrm{d}t} = \frac{\beta^2(1+\beta^2)\cdot P^2}{(\beta^2 \cdot P + M \cdot t)^2}.
\end{equation*}
It is strictly positive for all $t$ in its domain, thus $f$ is strictly increasing in $t$. This means $\EE[\FB]$ is non-decreasing in $\theta$ and also in $\THS$, because the term $\THS = \MT / M$ is non-decreasing in $\theta$.
Hence, the extreme values of the expectation of $\FB$ are its border values:
\begin{align*}
  \min_{\theta \in [1/(2M), 1]}\left(\EE[\FB]\right) & = \min_{\theta \in [1/(2M), 1]}\left(\frac{(1+\beta^2)\cdot P \cdot \MT}{M \cdot (\beta^2 \cdot P + \MT)} \right) \\
                                                     & = \frac{(1+\beta^2)\cdot P }{M(\beta^2\cdot P + 1)},                                                              \\
  \max_{\theta \in [1/(2M), 1]}\left(\EE[\FB]\right) & = \max_{\theta \in [1/(2M), 1]}\left(\frac{(1+\beta^2)\cdot P \cdot \MT}{M \cdot (\beta^2 \cdot P + \MT)} \right) \\
                                                     & = \frac{(1+\beta^2) \cdot P}{\beta^2 \cdot P + M}.
\end{align*}
Note that $\MT > 0$ is a restriction for $\FB$, and hence the optima are taken over the interval $[1/(2M), 1]$. Furthermore, the optimization values $\thmin$ and $\thmax$ for the extreme values are given by
\begin{align*}
  \thmin & \in \argmin_{\theta \in [1/(2M), 1]}\left(\EE[\FB]\right) = \argmin_{\theta \in [1/(2M), 1]}\left(\frac{\MT}{\beta^2 \cdot P + \MT}\right)       \\
         & \quad= \begin{cases}[\frac{1}{2}, 1] & \text{if $M = 1$} \\ \left[\frac{1}{2M}, \frac{3}{2M}\right) & \text{if $M > 1$,} \end{cases}
\end{align*}
\begin{align*}
  \thmax & \in \argmax_{\theta \in [1/(2M), 1]}\left(\EE[\FB]\right) = \argmax_{\theta \in [1/(2M), 1]}\left(\frac{\MT}{\beta^2 \cdot P + \MT}\right) \\
         & \quad = \left[1 - \frac{1}{2M}, 1\right],
\end{align*}
respectively. Following this reasoning, the discrete forms $\thsmin$ and $\thsmax$ are given by
\begin{align*}
  \thsmin & \in \argmin_{\THS \in \THSSPACE \setminus \{0\}}\left\{\EE[\FBS]\right\} = \argmin_{\THS \in \THSSPACE \setminus \{0\}}\left\{\frac{\THS}{\beta^2\cdot P + M\cdot\THS}\right\} = \left\{\frac{1}{M}\right\}, \\
  \thsmax & \in \argmax_{\THS \in \THSSPACE \setminus \{0\}}\left\{\EE[\FBS]\right\} = \argmax_{\THS \in \THSSPACE \setminus \{0\}}\left\{\frac{\THS}{\beta^2\cdot P + M\cdot\THS}\right\} = \{1\}.
\end{align*}
The smallest $\EE[\FB]$ is obtained when all observations except one are predicted negative, while predicting everything positive yields the largest $\EE[\FB]$. %

\begin{table}[H]
  \centering
  \def\arraystretch{1.8}
  \caption{\textbf{\emph{\FancyName{} baseline:}}
    For many evaluation measures, the minimum and maximum expected score of all allowed %
    \FancyName{} classifiers is determined, which is the \FancyName{} baseline. In this table, the baselines and the optimizing parameters are given. ``\xmark'' denotes that no closed-form expression was found.}
  \label{tab: Optimal baseline measures}

  \rowcolors{2}{white}{gray!25}
  \begin{tabular}{lllll} \toprule %

    \bf Measure                                      & $\bm \max\{\EE\}$                                    & $\bm \Theta^\star_{\text{max}} := \argmax\{\EE\}$                                                                                                                         & $\bm \min\{\EE\}$                                   & $\bm \Theta^\star_{\text{min}} := \argmin\{\EE\}$         \\ \midrule
    \belowrulesepcolor{gray!25}
    \hyperref[subsec: Appendix TP]{TP}               & $P$                                                  & $\{1\}$                                                                                                                                                                   & $0$                                                 & $\{0\}$                                                   \\
    \hyperref[subsec: Appendix TN]{TN}               & $\MP$                                                & $\{0\}$                                                                                                                                                                   & $0$                                                 & $\{1\}$                                                   \\
    \hyperref[subsec: Appendix FN]{FN}               & $P$                                                  & $\{0\}$                                                                                                                                                                   & $0$                                                 & $\{1\}$                                                   \\
    \hyperref[subsec: Appendix FP]{FP}               & $\MP$                                                & $\{1\}$                                                                                                                                                                   & $0$                                                 & $\{0\}$                                                   \\
    \hyperref[subsec: Appendix TPR]{TPR}             & $1$                                                  & $\{1\}$                                                                                                                                                                   & $0$                                                 & $\{0\}$                                                   \\
    \hyperref[subsec: Appendix TNR]{TNR}             & $ 1$                                                 & $\{0\}$                                                                                                                                                                   & $0$                                                 & $\{1\}$                                                   \\
    \hyperref[subsec: Appendix FNR]{FNR}             & $1$                                                  & $\{0\}$                                                                                                                                                                   & $0$                                                 & $\{1\}$                                                   \\
    \hyperref[subsec: Appendix FPR]{FPR}             & $ 1$                                                 & $\{1\}$                                                                                                                                                                   & $0$                                                 & $\{0\}$                                                   \\
    \hyperref[subsec: Appendix PPV]{PPV}             & $\frac{P}{M}$                                        & $\THSSPACE \setminus \{0\}$                                                                                                                                               & $\frac{P}{M} $                                      & $\THSSPACE \setminus \{0\}$                               \\
    \hyperref[subsec: Appendix NPV]{NPV}             & $1 - \frac{P}{M}$                                    & $\THSSPACE \setminus \{1\}$                                                                                                                                               & $1 - \frac{P}{M} $                                  & $\THSSPACE \setminus \{1\}$                               \\
    \hyperref[subsec: Appendix FDR]{FDR}             & $1 - \frac{P}{M}$                                    & $\THSSPACE \setminus \{0\}$                                                                                                                                               & $1 - \frac{P}{M} $                                  & $\THSSPACE \setminus \{0\}$                               \\
    \hyperref[subsec: Appendix FOR]{FOR}             & $\frac{P}{M}$                                        & $\THSSPACE \setminus \{1\}$                                                                                                                                               & $\frac{P}{M} $                                      & $\THSSPACE \setminus \{1\}$                               \\
    \hyperref[subsec: Appendix FBETA]{$F_{\beta}$}   & $\frac{(1+ \beta ^2) \cdot P}{\beta ^2 \cdot P + M}$ & $\{1\}$                                                                                                                                                                   & $\frac{(1+\beta^2)\cdot P }{M(\beta^2\cdot P + 1)}$ & $\left\{\frac{1}{M}\right\}$                              \\
    \hyperref[subsec: Appendix J]{J}                 & $ 0$                                                 & $\THSSPACE$                                                                                                                                                               & $0$                                                 & $\THSSPACE$                                               \\
    \hyperref[subsec: Appendix MK]{MK}               & $0$                                                  & $\THSSPACE \setminus \{0,1\}$                                                                                                                                             & $0$                                                 & $\THSSPACE \setminus \{0,1\}$                             \\
    \hyperref[subsec: Appendix ACC]{Acc}             & $\max \left\{\frac{P}{M}, 1-\frac{P}{M}\right\}$     & $\{[P < \frac{M}{2}] \} $ \tablefootnote{If $P = \frac{M}{2}$, then $\THSSPACE$. Note that Iverson brackets are used to simplify notation. \label{footnote: condition 1}} & $ \min \left\{\frac{P}{M}, 1-\frac{P}{M}\right\} $  & $\{[P > \frac{M}{2}] \} $ \footref{footnote: condition 1} \\
    \hyperref[subsec: Appendix BACC]{BAcc}           & $\frac{1}{2} $                                       & $\THSSPACE$                                                                                                                                                               & $\frac{1}{2} $                                      & $\THSSPACE$                                               \\
    \hyperref[subsec: Appendix MCC]{MCC}             & $0$                                                  & $\THSSPACE \setminus \{0,1\} $                                                                                                                                            & $0 $                                                & $\THSSPACE \setminus \{0,1\}$                             \\
    \hyperref[subsec: Appendix KAPPA]{$\kappa$}      & $0 $                                                 & $\THSSPACE$ \tablefootnote{If $P=M$, then $\THSSPACE \setminus \{1\}$. \label{footnote: condition 2}}                                                                     & $0$                                                 & $\THSSPACE$ \footref{footnote: condition 2}               \\
    \hyperref[subsec: Appendix FM]{FM}               & $\sqrt{\frac{P}{M}}$                                 & $\{1\}$                                                                                                                                                                   & $\frac{\sqrt{P}}{M} $                               & $\{\frac{1}{M}\}$                                         \\
    \hyperref[subsec: Appendix G2]{$\text{G}^{(2)}$} & \xmark                                               & \xmark                                                                                                                                                                    & $0$                                                 & $\{0, 1\}$                                                \\
    \hyperref[subsec: Appendix TS]{TS}               & $\frac{P}{M}$                                        & $\{1\}$                                                                                                                                                                   & $0$                                                 & $\{0\}$                                                   \\ \bottomrule
  \end{tabular}
  \vspace*{1 mm}

\end{table}

\section{Dutch Draw in Practice} \label{sec:experiments}

Now that we have established how to derive the DD baseline, it is time to see it in action. As a demonstration, we determined the DD baseline for commonly used evaluation measures on eight datasets extracted from the UCI machine learning archive~\citep{Dua2021}: \emph{Adult}, \emph{Bank Marketing}, \emph{Banknote Authentication}, \emph{Cleveland Heart Disease}, \emph{Haberman's Survival}, \emph{LSVT Voice Rehabilitation}, \emph{Occupancy Detection}, and \emph{Wisconsin Cancer}. The resulting DD baselines are shown in Table~\ref{tab: baselines_popular_datasets}. For some measures, the DD baseline already achieves the highest attainable score, such as for TPR and FNR. This suggests that these measures are not reliable indicators of the overall performance of a model. %
The problem is that these measures are only concerned with correctly predicting the positive instances. Always predicting positive therefore trivially gives the optimal performance. A less obvious DD baseline is the one for the $F_1$ score on the \emph{Bank Marketing} dataset. The DD achieves an expected performance of approximately $0.629$. Any new model for the \emph{Bank Marketing} dataset should therefore surpass this score. Next, we want to discuss what conclusions can be drawn from such a comparison by examining the following example.

\begin{table}[H]
  \centering
  \def\arraystretch{1.5}
  \caption{\textbf{\emph{\FancyName{} baseline for UCI datasets}:} Each dataset has different $P$ and $M$, resulting in different Dutch Draw baselines. }
  \label{tab: baselines_popular_datasets}

  \begin{adjustbox}{width=\linewidth, keepaspectratio}
    \rowcolors{2}{white}{gray!25}

    \begin{tabular}{lccccccccc} \toprule
      \bf Measure                                      & Adult \tablefootnote{\cite{Dua2021} \label{footnote: dataset 1}} & \thead{Bank                                                 \\ Marketing \tablefootnote{\cite{Moro2014}}} & \thead{Banknote \\ Authentication \footref{footnote: dataset 1}} & \thead{Cleveland Heart \\ Disease \footref{footnote: dataset 1}} & \thead{Haberman's \\ Survival \footref{footnote: dataset 1}} & \thead{LSVT Voice \\ Rehabilitation \tablefootnote{\cite{Tsanas2014}}} & \thead{Occupancy \\ Detection \tablefootnote{\cite{Candanedo2016}}} & \thead{Wisconsin Cancer \\ (Diagnostic) \footref{footnote: dataset 1}} \\
      \midrule
      \belowrulesepcolor{gray!25}
      \hyperref[subsec: Appendix TP]{TP}               & 11687                                                            & 5289        & 610   & 139   & 81    & 42    & 4750  & 212   \\
      \hyperref[subsec: Appendix TN]{TN}               & 37155                                                            & 39922       & 762   & 164   & 225   & 84    & 15810 & 357   \\
      \hyperref[subsec: Appendix FN]{FN}               & 37155                                                            & 39922       & 672   & 164   & 225   & 84    & 15810 & 357   \\
      \hyperref[subsec: Appendix FP]{FP}               & 11687                                                            & 5289        & 610   & 139   & 81    & 42    & 4750  & 212   \\
      \hyperref[subsec: Appendix TPR]{TPR}             & 1                                                                & 1           & 1     & 1     & 1     & 1     & 1     & 1     \\
      \hyperref[subsec: Appendix TNR]{TNR}             & 1                                                                & 1           & 1     & 1     & 1     & 1     & 1     & 1     \\
      \hyperref[subsec: Appendix FNR]{FNR}             & 1                                                                & 1           & 1     & 1     & 1     & 1     & 1     & 1     \\
      \hyperref[subsec: Appendix FPR]{FPR}             & 1                                                                & 1           & 1     & 1     & 1     & 1     & 1     & 1     \\
      \hyperref[subsec: Appendix PPV]{PPV}             & 0.239                                                            & 0.117       & 0.445 & 0.459 & 0.265 & 0.333 & 0.231 & 0.373 \\
      \hyperref[subsec: Appendix NPV]{NPV}             & 0.761                                                            & 0.883       & 0.555 & 0.541 & 0.735 & 0.667 & 0.769 & 0.627 \\
      \hyperref[subsec: Appendix FDR]{FDR}             & 0.761                                                            & 0.883       & 0.555 & 0.541 & 0.735 & 0.667 & 0.769 & 0.627 \\
      \hyperref[subsec: Appendix FOR]{FOR}             & 0.239                                                            & 0.117       & 0.445 & 0.459 & 0.265 & 0.333 & 0.231 & 0.373 \\
      \hyperref[subsec: Appendix FBETA]{$F_{1}$}       & 0.386                                                            & 0.209       & 0.616 & 0.629 & 0.419 & 0.5   & 0.375 & 0.543 \\
      \hyperref[subsec: Appendix J]{J}                 & 0                                                                & 0           & 0     & 0     & 0     & 0     & 0     & 0     \\
      \hyperref[subsec: Appendix MK]{MK}               & 0                                                                & 0           & 0     & 0     & 0     & 0     & 0     & 0     \\
      \hyperref[subsec: Appendix ACC]{Acc}             & 0.761                                                            & 0.883       & 0.555 & 0.541 & 0.735 & 0.667 & 0.769 & 0.627 \\
      \hyperref[subsec: Appendix BACC]{BAcc}           & 0.5                                                              & 0.5         & 0.5   & 0.5   & 0.5   & 0.5   & 0.5   & 0.5   \\
      \hyperref[subsec: Appendix MCC]{MCC}             & 0                                                                & 0           & 0     & 0     & 0     & 0     & 0     & 0     \\
      \hyperref[subsec: Appendix KAPPA]{$\kappa$}      & 0                                                                & 0           & 0     & 0     & 0     & 0     & 0     & 0     \\
      \hyperref[subsec: Appendix FM]{FM}               & 0.489                                                            & 0.342       & 0.667 & 0.677 & 0.514 & 0.577 & 0.481 & 0.61  \\
      \hyperref[subsec: Appendix G2]{$\text{G}^{(2)}$} & 0.5                                                              & 0.5         & 0.5   & 0.5   & 0.5   & 0.5   & 0.5   & 0.5   \\
      \hyperref[subsec: Appendix TS]{TS}               & 0.239                                                            & 0.117       & 0.445 & 0.459 & 0.265 & 0.333 & 0.231 & 0.373 \\ \bottomrule
    \end{tabular}
  \end{adjustbox}
  \vspace*{1 mm}

\end{table}

\subsection{Example: Cleveland Heart disease}
\begin{figure}[H]
  \centering
  \includegraphics[width=0.6\linewidth]{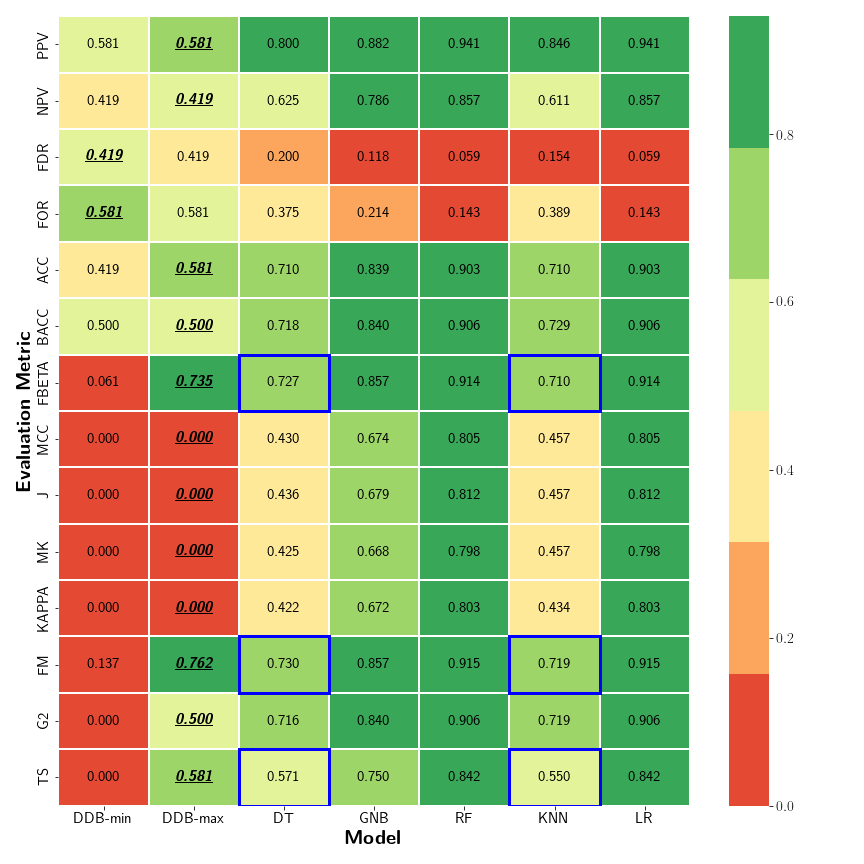}
  \caption{\textbf{Comparing performance to the DD baseline:}
    Five standard machine learning algorithms (\emph{logistic regression}, \emph{decision tree}, \emph{random forest}, \emph{$k$-nearest neighbors}, and \emph{gaussian naive bayes}) are tested on the \emph{Cleveland Heart Disease} dataset for many commonly used performance measures. The results are compared with the minimal and maximal DD baseline (DDB), which are given in the first two columns. Bold and underlined indicates that this score is more relevant, as the performance measure is commonly minimized or maximized. The blue boxes highlight some situations where a model achieves a score inferior to the DD baseline}
  \label{fig: heart comparison with baseline}
\end{figure}

The objective of this dataset is to predict whether patients have a heart disease given several feature values. In order to do so, we used five commonly used machine learning algorithms to perform this binary classification task: \emph{logistic regression}, \emph{decision tree}, \emph{random forest}, \emph{$k$-nearest neighbors}, and \emph{Gaussian naive Bayes}. These algorithms all had their default parameters in \emph{scikit-learn} \citep{scikit-learn}. The dataset was randomly split in a training (90\%) and test set (10\%). Fig.~\ref{fig: heart comparison with baseline} shows the corresponding performance results.

Before applying a newly developed model to actual patients, its performance should at least be better than the DD baseline, %
as the latter does not learn anything from the feature values of the data. In Fig.~\ref{fig: heart comparison with baseline}, we see that some methods fail to beat the baseline and should therefore be reconsidered. For example, \emph{decision tree} and \emph{$k$-nearest neighbors} underperform for the \emph{$F_{\beta}$ score} (FBETA), \emph{Fowlkes-Mallows Index} (FM), and \emph{Threat Score} (TS). %
Note that the two methods
were not trained to be optimal for the selected performance measures, whereas the DD does take the performance measure into account. However, this does not make the comparison unfair, since they are not competing for being the best prediction method. %
After all, the DD baseline is a minimal requirement for any new binary classification method. Even though a model is optimized for, say, the Accuracy, its performance should still beat the DD baseline for the $F_1$ score, as both the Accuracy and $F_1$ score provide indications of the overall prediction performance.
To conclude, this example shows how the DD can be used in practice and why it is valuable in the evaluation process.

\section{Discussion and Conclusion} \label{sec:discussion}
In this research, we have proposed a new baseline methodology called the \emph{Dutch Draw} (DD). The DD baseline is:
\begin{enumerate*}%
  \item [(1)] applicable to any binary classification problem;
  \item [(2)] reproducible;
  \item [(3)] simple;
  \item [(4)] parameter-free;
  \item [(5)] more informative than any single dummy baseline;
  \item [(6)] and an explainable minimal requirement for any new model.
\end{enumerate*}
We have shown that for most commonly used measures the DD baseline can be theoretically determined (see Table~\ref{tab: Optimal baseline measures}). When the baseline cannot be derived directly, it can be identified quickly by computation. For most performance measures, the DD baseline reduces to one of the following three cases: \begin{enumerate*}[label=(\roman*)]
  \item always predicting positive or negative;
  \item always predicting positive or negative, except for one instance;
  \item any DD classifier, except maybe for $\THS = 0$ or $\THS = 1$.
\end{enumerate*}
However, there are exceptions to these three cases. Examine the following example for the \emph{G-mean 2}: $P = 9$ and $M = 10$. We have previously seen (Table~\ref{tab: Optimal baseline measures}) that $\THS \in \{0,1\}$ achieves the \emph{lowest} expected score. To find the \emph{highest} expected score, note that $\THS = \frac{1}{M}$ gives an expected score of $\frac{3}{10}$, $\THS = \frac{2}{M}$ a score of $\frac{4\sqrt{2}}{15}$, and $\THS = \frac{M-1}{M}$ a score of $\frac{1}{10}$. This shows that the optimal parameter is not in $\{0, \frac{1}{M}, \frac{M-1}{M}, 1\}$, as $\frac{4\sqrt{2}}{15} > \frac{3}{10} > \frac{1}{10}$. In this case, the maximum is achieved for $\THS = \frac{3}{10}$. This shows that the DD does not always reduce to one of the three previously mentioned cases and does not always give straightforward results.

By introducing the DD baseline, we have simplified and improved the evaluation process of new binary classification methods. We consider it a minimal requirement for any novel model to at least beat the DD baseline. When this does not happen, the question is raised how much a new method has even learned from the data, since the DD baseline is derived from dummy classifiers. When the novel model has beaten the DD baseline, it should still be compared to a state-of-the-art method in that specific domain to obtain additional insights. %
In Sec.~\ref{sec:experiments}, we have shown how the DD should be used in practice and that commonly used approaches such as \emph{$k$-nearest neighbors} and a \emph{decision tree} can underperform. Hence, using the Dutch Draw as a general, simple and informative baseline should be the new gold standard in any model evaluation process.

\subsection{Further research}
Our baseline is a stepping stone for further research, where multiple avenues should be explored. We discuss four possible research directions.

Firstly, we are now able to determine whether a binary classification model performs better than a universal baseline. However, we do not yet know how \emph{much} it performs better (or worse). For example, let the baseline have a score of 0.5 and a new model a score of 0.9. How much better is the latter score? It could be that a tiny bit of extra information easily pushes the score from 0.5 to 0.9. Or, it is possible that a model needs a lot of information to understand the intricacies of the problem, making it very difficult to reach a score of 0.9. Thus, it is necessary to quantify how hard it is to reach any score. Also, when another model is added that achieves a score of 0.91, can the difference in performance of these models be quantified? Is it only a slightly better model or is it a leap forward?

Secondly, our \FancyName{} baseline could be used to construct new standardized evaluation measures from their original versions. The advantage of these new measures would be that the interpretation of their scores is independent of the number of positive and negative observations in the dataset. In other words, the \FancyName{} baseline would already be incorporated in the new measure, such that comparing a score to the baseline is not necessary anymore. There are many ways how the \FancyName{} baseline can be used to scale a measure. Let $\DDmax$ and $\DDmin$ denote the maximum and minimum Dutch Draw baseline, respectively. As an example, a measure $\mu$ with range $[\mu_{\min},\mu_{\max}]$ that needs to be maximized can be rescaled by
\begin{align*}
  \mu_{\text{rescaled}} & =
  \left \{ \begin{array}{ll}
             -1                                       & \text{if } \mu \leq \DDmin,              \\
             \frac{\mu - \DDmax}{\DDmax - \DDmin}     & \text{if } \DDmin \leq \mu \leq \DDmax , \\
             \frac{\mu - \DDmax}{\mu_{\max} - \DDmax} & \text{else}.
           \end{array} \right.
\end{align*}
Everything below the lowest Dutch Draw baseline ($\DDmin$) gets value $-1$, because every Dutch Draw classifier is then performing better. This should be a major warning sign. A score between $\DDmin$ and $\DDmax$ is rescaled to $[-1,0]$. This value indicates that the performance is still worse than the best Dutch Draw baseline. All scores above $\DDmax$ are scaled to $[0,1].$ In this case, the performance at least performed better than the best Dutch Draw baseline.

Thirdly, another natural extension would be to drop the binary assumption and consider multiclass classification. This is more complicated than it seems, because not every multiclass evaluation measure follows automatically from its binary counterpart. However, we expect that for most multiclass measures it is again optimal to always predict a single specific class.

Fourthly, the essence of the \FancyName{} could be used to create universal baselines for other prediction problems, such as for regression problems. This means an approach that also uses (almost) no information from the data and is able to generate a measure-specific baseline to which newly developed models could be compared.

As a final note, we have published the code for the \FancyName{}, such that the reader can easily implement the baseline into their binary classification problems.\footnote{\url{https://github.com/joris-pries/DutchDraw}}

\appendix

\section{Mathematical Derivations}
\label{app: Appendix}

This section contains the complete theoretical analysis that is used to gather the information presented in Sec.~\ref{sec:preliminaries} and~\ref{sec:methodology}, and more specifically, Table~\ref{tab: Assumptions P and N},~\ref{tab: Properties baseline measures} and~\ref{tab: Optimal baseline measures}. Each subsection is dedicated to one of the evaluation measures. The following definitions are frequently used throughout this section:
\begin{align*}
  \Zab{\theta}{a}{b} & := a\cdot\TP + b \text{ with } a, b \in \mathbb{R}          \\
  \Hab{\theta}{a}{b} & := \text{probability distribution of $\Zab{\theta}{a}{b}$}.
\end{align*}

An overview of the entire Appendix can be viewed in Table~\ref{tab: appendix}.

\begin{table}[H]
  \centering
  \def\arraystretch{1.6}
  \caption{\textbf{\emph{Overview of the Appendix:}}
    Each measure is discussed in the corresponding section in the Appendix}
  \label{tab: appendix}
  \begin{adjustbox}{width=\linewidth,keepaspectratio}
    \begin{tabular}{ccccccccccccc} \toprule
      {\bf Measure}                                    & \hyperref[subsec: Appendix TP]{TP}             &
      \hyperref[subsec: Appendix TN]{TN}               &
      \hyperref[subsec: Appendix FN]{FN}               &
      \hyperref[subsec: Appendix FP]{FP}               &
      \hyperref[subsec: Appendix TPR]{TPR}             &
      \hyperref[subsec: Appendix TNR]{TNR}             &
      \hyperref[subsec: Appendix FNR]{FNR}             &
      \hyperref[subsec: Appendix FPR]{FPR}             &
      \hyperref[subsec: Appendix PPV]{PPV}             &
      \hyperref[subsec: Appendix NPV]{NPV}             &
      \hyperref[subsec: Appendix FDR]{FDR}             &
      \hyperref[subsec: Appendix FOR]{FOR}                                                                \\
      {\bf Section}                                    & \ref{subsec: Appendix TP}                      &
      \ref{subsec: Appendix TN}                        &
      \ref{subsec: Appendix FN}                        &
      \ref{subsec: Appendix FP}                        &
      \ref{subsec: Appendix TPR}                       &
      \ref{subsec: Appendix TNR}                       &
      \ref{subsec: Appendix FNR}                       &
      \ref{subsec: Appendix FPR}                       &
      \ref{subsec: Appendix PPV}                       &
      \ref{subsec: Appendix NPV}                       &
      \ref{subsec: Appendix FDR}                       &
      \ref{subsec: Appendix FOR}                                                                          \\ \midrule
      {\bf Measure}                                    & \hyperref[subsec: Appendix FBETA]{$F_{\beta}$} &
      \hyperref[subsec: Appendix J]{J}                 &
      \hyperref[subsec: Appendix MK]{MK}               &
      \hyperref[subsec: Appendix ACC]{Acc}             &
      \hyperref[subsec: Appendix BACC]{BAcc}           &
      \hyperref[subsec: Appendix MCC]{MCC}             &
      \hyperref[subsec: Appendix KAPPA]{$\kappa$}      &
      \hyperref[subsec: Appendix FM]{FM}               &
      \hyperref[subsec: Appendix G2]{$\text{G}^{(2)}$} &
      \hyperref[subsec: Appendix PT]{PT}               &
      \hyperref[subsec: Appendix TS]{TS}                                                                  \\
      {\bf Section}                                    & \ref{subsec: Appendix FBETA}                   &
      \ref{subsec: Appendix J}                         &
      \ref{subsec: Appendix MK}                        &
      \ref{subsec: Appendix ACC}                       &
      \ref{subsec: Appendix BACC}                      &
      \ref{subsec: Appendix MCC}                       &
      \ref{subsec: Appendix KAPPA}                     &
      \ref{subsec: Appendix FM}                        &
      \ref{subsec: Appendix G2}                        &
      \ref{subsec: Appendix PT}                        &
      \ref{subsec: Appendix TS}
      \\ \bottomrule
    \end{tabular}
  \end{adjustbox}
  \vspace*{1 mm}

\end{table}

\subsection{Number of True Positives}
\label{subsec: Appendix TP}

The \textit{Number of True Positives} $\TP$ is one of the four base measures that are introduced in Sec.~\ref{subsec:evaluation_measures}. This measure indicates how many of the predicted positive observations are actually positive. Under the \FancyName{} methodology, each evaluation measure can be written in terms of $\TP$.

\subsubsection{Definition and Distribution}

Since we want to formulate each measure in terms of $\TP$, we have for $\TP$:
\begin{equation*}
  \TP \stackrel{\eqref{eq: TP in TP}}{=} \Zab{\theta}{1}{0} \sim \Hab{\theta}{1}{0}.
\end{equation*}
The range of this base measure depends on $\theta$. Therefore, Eq.~\eqref{eq: Range Zab} yields the range of this measure:
\begin{equation*}
  \TP \in \RZab{\theta}{1}{0}.
\end{equation*}

\subsubsection{Expectation}

The expectation of $\TP$ using the \FancyName{} is given by
\begin{align}
  \EE[\TP] & = \EE[\Zab{\theta}{1}{0}] \stackrel{\eqref{eq: Expectation Rule 1}}{=} \frac{\MT}{M} \cdot P = \THS \cdot P. \label{eq:apptpe}
\end{align}

\subsubsection{Optimal Baselines}

The \FancyName{} baseline is given by the optimal expectation. Eq.~\eqref{eq:apptpe} shows that the expected value depends on the parameter $\theta$. Therefore, either the minimum or maximum of the expectation yields the baseline. They are given by
\begin{align*}
  \THMIN{\EE[\TP]} & = P \cdot \THMIN{\frac{\MT}{M}} = 0, \\
  \THMAX{\EE[\TP]} & = P \cdot \THMAX{\frac{\MT}{M}} = P.
\end{align*}
The values of $\theta \in [0,1]$ that minimize or maximize the expected value are $\thmin$ and $\thmax$, respectively, and are defined as
\begin{align*}
  \thmin & \in \THARGMIN{\EE[\TP]} = \THARGMIN{\frac{\MT}{M}} = \left[0, \frac{1}{2M}\right),     \\
  \thmax & \in \THARGMAX{\EE[\TP]} = \THARGMAX{\frac{\MT}{M}} = \left[1 - \frac{1}{2M}, 1\right].
\end{align*}
Equivalently, the discrete optimizers $\thsmin \in \THSSPACE$ and $\thsmax \in \THSSPACE$ are determined by
\begin{align*}
  \thsmin & \in \THSARGMIN{\EE[\TPS]} = \THSARGMIN{\THS} = \{0\}, \\
  \thsmax & \in \THSARGMAX{\EE[\TPS]} = \THSARGMAX{\THS} = \{1\}.
\end{align*}

\subsection{Number of True Negatives}\label{subsec: Appendix TN}

The \textit{Number of True Negatives} $\TN$ is also one of the four base measures and is introduced in Sec.~\ref{subsec:evaluation_measures}. This base measure counts the number of negative predicted instances that are actually negative.

\subsubsection{Definition and Distribution}

Since we want to formulate each measure in terms of $\TP$, we have for $\TN$:
\begin{align*}
  \TN & = \MP - \MT + \TP,
\end{align*}
which corresponds to Eq.~\eqref{eq: TN in TP}. Furthermore,
\begin{equation*}
  \TN \stackrel{\eqref{eq: TN in TP}}{=} \Zab{\theta}{1}{\MP - \MT} \sim \Hab{\theta}{1}{\MP - \MT},
\end{equation*}
and for its range
\begin{align*}
  \TN \stackrel{\eqref{eq: Range Zab}}{\in} \RZab{\theta}{1}{\MP - \MT}.
\end{align*}

\subsubsection{Expectation}

$\TN$ is linear in $\TP$ with slope $a = 1$ and intercept $b = \MP - \MT$, so its expectation is given by
\begin{align*}
  \EE[\TN] & = \EE[\Zab{\theta}{1}{\MP - \MT}] \stackrel{\eqref{eq: Expectation Rule 1}}{=} 1 \cdot \EE[\TP] + \MP-\MT \\
           & = \left(1 - \frac{\MT}{M}\right)\MPP = (1-\THS)\MPP.
\end{align*}

\subsubsection{Optimal Baselines}

To determine the range of the expectation of $\TN$, and hence, obtain baselines, its extreme values are calculated:
\begin{align*}
  \THMIN{\EE[\TN]} & = \MPP \THMIN{1 - \frac{\MT}{M}} = 0,   \\
  \THMAX{\EE[\TN]} & = \MPP \THMAX{1 - \frac{\MT}{M}} = \MP.
\end{align*}
The associated optimization values $\thmin \in [0,1]$ and $\thmax \in [0,1]$ are
\begin{align*}
  \thmin & \in \THARGMIN{\EE[\TN]} = \THARGMIN{1 - \frac{\MT}{M}} = \left[1 - \frac{1}{2M}, 1\right], \\
  \thmax & \in \THARGMAX{\EE[\TN]} = \THARGMAX{1 - \frac{\MT}{M}} = \left[0, \frac{1}{2M}\right).
\end{align*}
The discrete equivalents $\thsmin \in \THSSPACE$ and $\thsmax \in \THSSPACE$ are then determined by
\begin{align*}
  \thsmin & \in \THSARGMIN{\EE[\TNS]} = \THSARGMIN{1-\THS} = \{1\}, \\
  \thsmax & \in \THSARGMAX{\EE[\TNS]} = \THSARGMAX{1-\THS} = \{0\}.
\end{align*}

\subsection{Number of False Negatives}\label{subsec: Appendix FN}

The \textit{Number of False Negative} $\FN$ is one of the four base measures that are introduced in Sec.~\ref{subsec:evaluation_measures}. This base measure counts the number of mistakes made by predicting instances negative while the actual labels are positive.

\subsubsection{Definition and Distribution}

Eq.~\eqref{eq: FN in TP} shows that $\FN$ can be expressed in terms of $\TP$:
\begin{equation*}
  \FN \stackrel{\eqref{eq: FN in TP}}{=} P - \TP = \Zab{\theta}{-1}{P} \sim \Hab{\theta}{-1}{P},
\end{equation*}
and for its range:
\begin{align*}
  \FN \stackrel{\eqref{eq: Range Zab}}{\in} \RZab{\theta}{-1}{P}.
\end{align*}

\subsubsection{Expectation}

As Eq.~\eqref{eq: FN in TP} shows, $\FN$ is linear in $\TP$ with slope $a = -1$ and intercept $b = P$. Hence, the expectation of $\FN$ is given by
\begin{align*}
  \EE[\FN] & = \EE[\Zab{\theta}{-1}{P}] \stackrel{\eqref{eq: Expectation Rule 1}}{=} -1 \cdot \EE[\TP] + P = \left(1-\frac{\MT}{M}\right)\cdot P = \left(1-\THS\right) \cdot P.
\end{align*}

\subsubsection{Optimal Baselines}

The range of the expectation of $\FN$ determines the baselines. The extreme values are given by
\begin{align*}
  \THMIN{\EE[\FN]} & = P \cdot \THMIN{1-\frac{\MT}{M}} = 0, \\
  \THMAX{\EE[\FN]} & = P \cdot \THMAX{1-\frac{\MT}{M}} = P.
\end{align*}
The associated optimization values $\thmin \in [0,1]$ and $\thmax \in [0,1]$ are then
\begin{align*}
  \thmin & \in \THARGMIN{\EE[\FN]} = \THARGMIN{1-\frac{\MT}{M}} = \left[1 - \frac{1}{2M}, 1\right], \\
  \thmax & \in \THARGMAX{\EE[\FN]} = \THARGMAX{1-\frac{\MT}{M}} = \left[0, \frac{1}{2M}\right),
\end{align*}
respectively.
The discrete versions $\thsmin \in \THSSPACE$ and $\thsmax \in \THSSPACE$ of the optimizers are as follows:
\begin{align*}
  \thsmin & \in \THSARGMIN{\EE[\FNS]} = \THSARGMIN{1-\THS} = \{1\}, \\
  \thsmax & \in \THSARGMAX{\EE[\FNS]} = \THSARGMAX{1-\THS} = \{0\}.
\end{align*}

\subsection{Number of False Positives}
\label{subsec: Appendix FP}

The \textit{Number of False Positives} $\FP$ is one of the four base measures that we discussed in Sec.~\ref{subsec:evaluation_measures}. This base measure counts the number of mistakes made by predicting instances positive while the actual labels are negative.

\subsubsection{Definition and Distribution}

Each base measure can be expressed in terms of $\TP$, thus we have for $\FP$:
\begin{equation*}
  \FP \stackrel{\eqref{eq: FP in TP}}{=} \MT - \TP = \Zab{\theta}{-1}{\MT}  \sim \Hab{\theta}{-1}{\MT},
\end{equation*}
and for its range:
\begin{align*}
  \FP \stackrel{\eqref{eq: Range Zab}}{\in} \RZab{\theta}{-1}{\MT}.
\end{align*}

\subsubsection{Expectation}

As Eq.~\eqref{eq: FP in TP} shows, $\FP$ is linear in $\TP$ with slope $a = -1$ and intercept $b = \MT$, thus the expectation of $\FP$ is defined as
\begin{align*}
  \EE[\FP] & = \EE[\Zab{\theta}{-1}{\MT}] \stackrel{\eqref{eq: Expectation Rule 1}}{=} -1 \cdot \EE[\TP] + \MT = \frac{\MT}{M}\cdot\MPP = \THS\cdot\MPP.
\end{align*}

\subsubsection{Optimal Baselines}

The baselines of $\FP$ are given by the extreme values of its expectation. Hence:
\begin{align*}
  \THMIN{\EE[\FP]} & = \MPP \THMIN{\frac{\MT}{M}} = 0,   \\
  \THMAX{\EE[\FP]} & = \MPP \THMAX{\frac{\MT}{M}} = \MP.
\end{align*}
The corresponding optimization values $\thmin \in [0,1]$ and $\thmax \in [0,1]$ are
\begin{align*}
  \thmin & \in \THARGMIN{\EE[\FP]} = \THARGMIN{\frac{\MT}{M}} = \left[0, \frac{1}{2M}\right),     \\
  \thmax & \in \THARGMAX{\EE[\FP]} = \THARGMAX{\frac{\MT}{M}} = \left[1 - \frac{1}{2M}, 1\right].
\end{align*}
The discrete versions $\thsmin \in \THSSPACE$ and $\thsmax \in \THSSPACE$ of the optimization values are determined by
\begin{align*}
  \thsmin & \in \THSARGMIN{\EE[\FPS]} = \THSARGMIN{\THS} = \{0\}, \\
  \thsmax & \in \THSARGMAX{\EE[\FPS]} = \THSARGMAX{\THS} = \{1\}.
\end{align*}

\subsection{True Positive Rate} \label{subsec: Appendix TPR}

The \textit{True Positive Rate} $\TPR$, \textit{Recall}, or \textit{Sensitivity} is the performance measure that presents the fraction of positive observations that are correctly predicted. This makes it a fundamental performance measure in binary classification.

\subsubsection{Definition and Distribution}

The True Positive Rate is commonly defined as
\begin{equation}
  \TPR = \frac{\TP}{P}. \label{eq:rawtprdef}
\end{equation}
Hence, $P > 0$ should hold, otherwise the denominator is zero. Now, $\TPR$ is linear in $\TP$ and can therefore be written as
\begin{equation}
  \TPR = \Zab{\theta}{\frac{1}{P}}{0} \sim \Hab{\theta}{\frac{1}{P}}{0},
  \label{eq:tprdef}
\end{equation}
and for its range:
\begin{align*}
  \TPR \stackrel{\eqref{eq: Range Zab}}{\in} \RZab{\theta}{\frac{1}{P}}{0}.
\end{align*}

\subsubsection{Expectation}

Since $\TPR$ is linear in $\TP$ with slope $a = 1/P$ and intercept $b = 0$, its expectation is
\begin{align*}
  \EE[\TPR] & = \EE\left[\Zab{\theta}{\frac{1}{P}}{0}\right] \stackrel{\eqref{eq: Expectation Rule 1}}{=} \frac{1}{P} \cdot \EE[\TP] + 0 = \frac{\MT}{M} = \THS.
\end{align*}

\subsubsection{Optimal Baselines}

The range of the expectation of $\TPR$ directly determines the baselines. The extreme values are given by
\begin{align*}
  \THMIN{\EE[\TPR]} & = \THMIN{\frac{\MT}{M}} = 0, \\
  \THMAX{\EE[\TPR]} & = \THMAX{\frac{\MT}{M}} = 1.
\end{align*}
Furthermore, the corresponding optimization values $\thmin \in [0,1]$ and $\thmax \in [0,1]$ are given by
\begin{align*}
  \thmin & \in \THARGMIN{\EE[\TPR]} = \THARGMIN{\frac{\MT}{M}} = \left[0, \frac{1}{2M}\right),     \\
  \thmax & \in \THARGMAX{\EE[\TPR]} = \THARGMAX{\frac{\MT}{M}} = \left[1 - \frac{1}{2M}, 1\right].
\end{align*}
The discrete versions $\thsmin \in \THSSPACE$ and $\thsmax \in \THSSPACE$ of the optimizers are then
\begin{align*}
  \thsmin & \in \THSARGMIN{\EE[\TPRS]} = \THSARGMIN{\THS} = \{0\}, \\
  \thsmax & \in \THSARGMAX{\EE[\TPRS]} = \THSARGMAX{\THS} = \{1\},
\end{align*}
respectively.

\subsection{True Negative Rate} \label{subsec: Appendix TNR}

The \textit{True Negative Rate} $\TNR$, \textit{Specificity}, or \textit{Selectivity} is the measure that shows how relatively well the negative observations are correctly predicted. Hence, this performance measure is a fundamental measure in binary classification.

\subsubsection{Definition and Distribution}

The True Negative Rate is commonly defined as
\begin{equation*}
  \TNR = \frac{\TN}{N}.
\end{equation*}
Hence, $N := M-P > 0$ should hold, otherwise the denominator is zero. By using Eq.~\eqref{eq: TN in TP}, $\TNR$ can be rewritten as
\begin{align*}
  \TNR = \frac{\MP - \MT + \TP}{\MP} = 1 - \frac{\MT - \TP}{\MP}.
\end{align*}
Hence, it is linear in $\TP$ and can therefore be written as
\begin{equation}
  \TNR = \Zab{\theta}{\frac{1}{\MP}}{1 - \frac{\MT}{\MP}} \sim \Hab{\theta}{\frac{1}{\MP}}{1 - \frac{\MT}{\MP}},
  \label{eq:tnrdef}
\end{equation}
and for its range:
\begin{align*}
  \TNR \stackrel{\eqref{eq: Range Zab}}{\in} \RZab{\theta}{\frac{1}{\MP}}{1 - \frac{\MT}{\MP}}.
\end{align*}

\subsubsection{Expectation}

Since $\TNR$ is linear in $\TP$ in terms of $\Zab{\theta}{a}{b}$ with slope $a = 1/\MPP$ and intercept $b = 1 - \MT/\MPP$, its expectation is
\begin{align*}
  \EE[\TNR] & = \EE\left[\Zab{\theta}{\frac{1}{\MP}}{1 - \frac{\MT}{\MP}}\right] \stackrel{\eqref{eq: Expectation Rule 1}}{=} \frac{1}{\MP} \cdot \EE[\TP] + 1 - \frac{\MT}{\MP} = 1 - \frac{\MT}{M} = 1 - \THS.
\end{align*}

\subsubsection{Optimal Baselines}

The extreme values of the expectation of $\TNR$ determine the baselines. The range is given by
\begin{align*}
  \THMIN{\EE[\TNR]} & = \THMIN{1-\frac{\MT}{M}} = 0, \\
  \THMAX{\EE[\TNR]} & = \THMAX{1-\frac{\MT}{M}} = 1.
\end{align*}
Moreover, the optimization values $\thmin \in [0,1]$ and $\thmax \in [0,1]$ corresponding to the extreme values are defined as
\begin{align*}
  \thmin & \in \THARGMIN{\EE[\TNR]} = \THARGMIN{1-\frac{\MT}{M}} = \left[1 - \frac{1}{2M}, 1\right], \\
  \thmax & \in \THARGMAX{\EE[\TNR]} = \THARGMAX{1-\frac{\MT}{M}} = \left[0, \frac{1}{2M}\right),
\end{align*}
respectively.
The discrete versions $\thsmin \in \THSSPACE$ and $\thsmax \in \THSSPACE$ of the optimizers are given by
\begin{align*}
  \thsmin & \in \THSARGMIN{\EE[\TNRS]} = \THSARGMIN{1-\THS} = \{1\}, \\
  \thsmax & \in \THSARGMAX{\EE[\TNRS]} = \THSARGMAX{1-\THS} = \{0\}.
\end{align*}

\subsection{False Negative Rate} \label{subsec: Appendix FNR}

The \textit{False Negative Rate} $\FNR$ or \textit{Miss Rate} is the performance measure that indicates the relative number of incorrectly predicted positive observations. Therefore, it can be seen as the counterpart to the True Positive Rate that is discussed in Sec.~\ref{subsec: Appendix TPR}.

\subsubsection{Definition and Distribution}

The False Negative Rate is commonly defined as
\begin{equation*}
  \FNR = \frac{\FN}{P}.
\end{equation*}
Hence, $P > 0$ should hold, otherwise the denominator is zero. With the aid of Eq.~\eqref{eq: FN in TP}, $\FNR$ can be reformulated to
\begin{align*}
  \FNR = \frac{P - \TP}{P} = 1 - \frac{\TP}{P}.
\end{align*}
Thus, it is linear in $\TP$ and can therefore be written as
\begin{equation*}
  \FNR = \Zab{\theta}{-\frac{1}{P}}{1} \sim \Hab{\theta}{-\frac{1}{P}}{1},
\end{equation*}
and for its range:
\begin{align*}
  \FNR \stackrel{\eqref{eq: Range Zab}}{\in} \RZab{\theta}{-\frac{1}{P}}{1}.
\end{align*}

\subsubsection{Expectation}

Because $\FNR$ is linear in $\TP$ with slope $a = -1/P$ and intercept $b = 1$, its expectation is
\begin{align*}
  \EE[\FNR] & = \EE\left[\Zab{\theta}{-\frac{1}{P}}{1}\right] \stackrel{\eqref{eq: Expectation Rule 1}}{=} -\frac{1}{P} \cdot \EE[\TP] + 1 = 1 - \frac{\MT}{M} = 1 - \THS.
\end{align*}

\subsubsection{Optimal Baselines}

The range of the expectation of $\FNR$ determines the baselines. The extreme values are given by:
\begin{align*}
  \THMIN{\EE[\FNR]} & = \THMIN{1-\frac{\MT}{M}} = 0, \\
  \THMAX{\EE[\FNR]} & = \THMAX{1-\frac{\MT}{M}} = 1.
\end{align*}
Furthermore, the optimizers $\thmin \in [0,1]$ and $\thmax \in [0,1]$ for the extreme values are as follows:
\begin{align*}
  \thmin & \in \THARGMIN{\EE[\FNR]} = \THARGMIN{1-\frac{\MT}{M}} = \left[1 - \frac{1}{2M}, 1\right], \\
  \thmax & \in \THARGMAX{\EE[\FNR]} = \THARGMAX{1-\frac{\MT}{M}} = \left[0, \frac{1}{2M}\right),
\end{align*}
respectively.
The discrete versions $\thsmin \in \THSSPACE$ and $\thsmax \in \THSSPACE$ of the optimization values are then:
\begin{align*}
  \thsmin & \in \THSARGMIN{\EE[\FNRS]} = \THSARGMIN{1-\THS} = \{1\}, \\
  \thsmax & \in \THSARGMAX{\EE[\FNRS]} = \THSARGMAX{1-\THS} = \{0\}.
\end{align*}

\subsection{False Positive Rate}\label{subsec: Appendix FPR}

The \textit{False Positive Rate} $\FPR$ or \textit{Fall-out} is the performance measure that shows the fraction of incorrectly predicted negative observations. Hence, it can be seen as the counterpart to the True Negative Rate that is introduced in Sec.~\ref{subsec: Appendix TNR}.

\subsubsection{Definition and Distribution}

The False Positive Rate is commonly defined as
\begin{equation*}
  \FPR = \frac{\FP}{N}.
\end{equation*}
Hence, $N := M-P$ should hold, otherwise the denominator is zero. By using Eq.~\eqref{eq: FP in TP}, $\FPR$ can be restated as
\begin{align}
  \FPR = \frac{\MT - \TP}{\MP}.
  \label{eq:fprdef}
\end{align}
Note that it is linear in $\TP$ and can therefore be written as
\begin{equation*}
  \FPR = \Zab{\theta}{-\frac{1}{\MP}}{\frac{\MT}{\MP}} \sim \Hab{\theta}{-\frac{1}{\MP}}{\frac{\MT}{\MP}},
\end{equation*}
with range:
\begin{align*}
  \FPR \stackrel{\eqref{eq: Range Zab}}{\in} \RZab{\theta}{-\frac{1}{\MP}}{\frac{\MT}{\MP}}.
\end{align*}

\subsubsection{Expectation}

Since $\FPR$ is linear in $\TP$ with slope $a =  -1/\MPP$ and intercept $b = \MT/\MPP$, its expectation is given by
\begin{align*}
  \EE[\FPR] & = \EE\left[\Zab{\theta}{-\frac{1}{\MP}}{\frac{\MT}{\MP}}\right] \stackrel{\eqref{eq: Expectation Rule 1}}{=} -\frac{1}{\MP} \cdot \EE[\TP] + \frac{\MT}{\MP} = \frac{\MT}{M} = \THS.
\end{align*}

\subsubsection{Optimal Baselines}

The extreme values of the expectation of $\FPR$ determine the baselines. The range is given by
\begin{align*}
  \THMIN{\EE[\FPR]} & = \THMIN{\frac{\MT}{M}} = 0, \\
  \THMAX{\EE[\FPR]} & = \THMAX{\frac{\MT}{M}} = 1.
\end{align*}
Moreover, the optimizers $\thmin \in [0,1]$ and $\thmax \in [0,1]$ for the extreme values are determined by
\begin{align*}
  \thmin & \in \THARGMIN{\EE[\FPR]} = \THARGMIN{\frac{\MT}{M}} = \left[0, \frac{1}{2M}\right),     \\
  \thmax & \in \THARGMAX{\EE[\FPR]} = \THARGMAX{\frac{\MT}{M}} = \left[1 - \frac{1}{2M}, 1\right],
\end{align*}
respectively.
The discrete forms $\thsmin \in \THSSPACE$ and $\thsmax \in \THSSPACE$ of these are then
\begin{align*}
  \thsmin & \in \THSARGMIN{\EE[\FPRS]} = \THSARGMIN{\THS} = \{0\}, \\
  \thsmax & \in \THSARGMAX{\EE[\FPRS]} = \THSARGMAX{\THS} = \{1\}.
\end{align*}

\subsection{Positive Predictive Value} \label{subsec: Appendix PPV}

The \textit{Positive Predictive Value} $\PPV$ or \textit{Precision} is the performance measure that considers the fraction of all positively predicted observations that are in fact positive. Therefore, it provides an indication of how cautious the model is in assigning positive predictions. A large value means the model is cautious in predicting observations as positive, while a small value means the opposite.

\subsubsection{Definition and Distribution}

The Positive Predictive Value is commonly defined as
\begin{equation}
  \PPV = \frac{\TP}{\TP + \FP}.
  \label{eq:rawppvdef}
\end{equation}
By using Eq.~\eqref{eq: TP in TP} and~\eqref{eq: FP in TP}, this definition can be reformulated to
\begin{equation*}
  \PPV = \frac{\TP}{\MT}.
\end{equation*}
Note that this performance measure is only defined whenever $\MT > 0$, otherwise the denominator is zero. Therefore, we assume specifically for $\PPV$ that $\theta \geq \frac{1}{2M}$.
The definition of $\PPV$ is linear in $\TP$ and can thus be formulated as
\begin{equation}
  \PPV = \Zab{\theta}{\frac{1}{\MT}}{0} \sim \Hab{\theta}{\frac{1}{\MT}}{0},
  \label{eq:ppvdef}
\end{equation}
with range:
\begin{align*}
  \PPV \stackrel{\eqref{eq: Range Zab}}{\in} \RZab{\theta}{\frac{1}{\MT}}{0}.
\end{align*}

\subsubsection{Expectation}

Because $\PPV$ is linear in $\TP$ with slope $a = 1/\MT$ and intercept $b = 0$, its expectation is
\begin{align*}
  \EE[\PPV] & = \EE\left[\Zab{\theta}{\frac{1}{\MT}}{0}\right] \stackrel{\eqref{eq: Expectation Rule 1}}{=} \frac{1}{\MT} \cdot \EE[\TP] + 0 = \frac{P}{M}.
\end{align*}

\subsubsection{Optimal Baselines}

The baselines are determined by the extreme values of the expectation of $\PPV$:
\begin{align*}
  \min_{\theta \in [1/(2M), 1]} \left(\EE[\PPV]\right) & = \frac{P}{M}, \\
  \max_{\theta \in [1/(2M), 1]} \left(\EE[\PPV]\right) & = \frac{P}{M},
\end{align*}
because the expectation does not depend on $\theta$.
Hence, the optimization values $\thmin$ and $\thmax$ are simply all allowed values for $\theta$:
\begin{equation*}
  \thmin = \thmax \in \left[\frac{1}{2M}, 1\right].
\end{equation*}
Consequently, the discrete versions $\thsmin$ and $\thsmax$ of these optimizers are in the set of all allowed discrete values:
\begin{equation*}
  \thsmin = \thsmax \in \THSSPACE \setminus \{0\}.
\end{equation*}

\subsection{Negative Predictive Value} \label{subsec: Appendix NPV}

The \textit{Negative Predictive Value} $\NPV$ is the performance measure that indicates the fraction of all negatively predicted observations that are in fact negative. Hence, it shows how cautious the model is in assigning negative predictions. A large value means the model is cautious in predicting observations negatively, while a small value means the opposite.

\subsubsection{Definition and Distribution}

The \textit{Negative Predictive Value} is commonly defined as
\begin{equation*}
  \NPV = \frac{\TN}{\TN + \FN}.
\end{equation*}
With the help of Eq.~\eqref{eq: FN in TP} and~\eqref{eq: TN in TP}, this definition can be rewritten as
\begin{align*}
  \NPV = 1 - \frac{P - \TP}{M - \MT}.
\end{align*}
Note that this performance measure is only defined whenever $\MT < M$, otherwise the denominator is zero. Therefore, we assume specifically for $\NPV$ that $\theta < 1 - \frac{1}{2M}$.
The definition of $\NPV$ is linear in $\TP$ and can thus be formulated as
\begin{align}
  \NPV & = \Zab{\theta}{\frac{1}{M-\MT}}{1 - \frac{P}{M - \MT}} \sim \Hab{\theta}{\frac{1}{M-\MT}}{1 - \frac{P}{M - \MT}}, \label{eq:npvdef}
\end{align}
with range:
\begin{align*}
  \NPV \stackrel{\eqref{eq: Range Zab}}{\in} \RZab{\theta}{\frac{1}{M-\MT}}{1 - \frac{P}{M - \MT}}.
\end{align*}

\subsubsection{Expectation}

Since $\NPV$ is linear in $\TP$ with slope $a = 1/(M-\MT)$ and intercept $b = 1 - P/(M-\MT)$, its expectation is given by
\begin{align*}
  \EE[\NPV] & = \EE\left[\Zab{\theta}{\frac{1}{M-\MT}}{1 - \frac{P}{M - \MT}}\right] \stackrel{\eqref{eq: Expectation Rule 1}}{=} \frac{1}{M-\MT} \cdot \EE[\TP] + 1 - \frac{P}{M - \MT} = 1 - \frac{P}{M}.
\end{align*}

\subsubsection{Optimal Baselines}

The extreme values of the expectation of $\NPV$ determine the baselines. They are given by
\begin{align*}
  \min_{\theta \in [0, 1 - 1/(2M))} \left(\EE[\NPV]\right) & = 1 - \frac{P}{M}, \\
  \max_{\theta \in [0, 1 - 1/(2M))} \left(\EE[\NPV]\right) & = 1 - \frac{P}{M},
\end{align*}
because the expectation does not depend on $\theta$.
Consequently, the optimization values $\thmin$ and $\thmax$ are all allowed values for $\theta$:
\begin{equation*}
  \thmin = \thmax \in \left[0, 1 - \frac{1}{2M}\right).
\end{equation*}
This also means the discrete forms $\thsmin$ and $\thsmax$ of the optimizers are in the set of all allowed discrete values:
\begin{equation*}
  \thsmin = \thsmax \in \THSSPACE \setminus \{1\}.
\end{equation*}

\subsection{False Discovery Rate} \label{subsec: Appendix FDR}

The \textit{False Discovery Rate} $\FDR$ is the performance measure that looks at the fraction of positively predicted observations that are actually negative. Therefore, it can be seen as the counterpart to the Positive Predictive Value that we discuss in Sec.~\ref{subsec: Appendix PPV}. Consequently, a small value means the model is cautious in predicting observations as positive, while a large value means the opposite.

\subsubsection{Definition and Distribution}

The \textit{False Discovery Rate} is commonly defined as
\begin{align*}
  \FDR & = \frac{\FP}{\TP + \FP} = 1 - \PPV.
\end{align*}
With the help of Eq.~\eqref{eq:ppvdef}, this definition can be rewritten as
\begin{align*}
  \FDR & = 1 - \frac{\TP}{\MT}.
\end{align*}
Note that this performance measure is only defined whenever $\MT > 0$, otherwise the denominator is zero. Therefore, we assume specifically for $\FDR$ that $\theta > \frac{1}{2M}$. The definition of $\FDR$ is linear in $\TP$ and can thus be formulated as
\begin{align*}
  \FDR & = \Zab{\theta}{-\frac{1}{\MT}}{1} \sim \Hab{\theta}{-\frac{1}{\MT}}{1},
\end{align*}
with range:
\begin{align*}
  \FDR \stackrel{\eqref{eq: Range Zab}}{\in} \RZab{\theta}{-\frac{1}{\MT}}{1}.
\end{align*}

\subsubsection{Expectation}

Since $\FDR$ is linear in $\TP$ with slope $a = -1/\MT$ and intercept $b = 1$, its expectation is given by
\begin{align*}
  \EE[\FDR] & = \EE\left[\Zab{\theta}{-\frac{1}{\MT}}{1}\right] \stackrel{\eqref{eq: Expectation Rule 1}}{=} -\frac{1}{\MT} \cdot \EE[\TP] + 1 = 1 - \frac{P}{M}.
\end{align*}

\subsubsection{Optimal Baselines}

The extreme values of the expectation of $\FDR$ determine the baselines. Its range is given by
\begin{align*}
  \min_{\theta \in (1/(2M),1]} \left(\EE[\FDR]\right) & = 1 - \frac{P}{M}, \\
  \max_{\theta \in (1/(2M),1]} \left(\EE[\FDR]\right) & = 1 - \frac{P}{M},
\end{align*}
because the expectation does not depend on $\theta$.
Consequently, the optimization values $\thmin$ and $\thmax$ are all allowed values for $\theta$:
\begin{equation*}
  \thmin = \thmax \in \left(\frac{1}{2M},1\right].
\end{equation*}
This also means the discrete forms $\thsmin$ and $\thsmax$ of the optimizers are in the set of all allowed discrete values:
\begin{equation*}
  \thsmin = \thsmax \in \THSSPACE \setminus \{0\}.
\end{equation*}

\subsection{False Omission Rate} \label{subsec: Appendix FOR}

The \textit{False Omission Rate} $\FOR$ is the performance measure that considers the fraction of observations that are predicted negative, but are in fact positive. Hence, it can be seen as the counterpart to the Negative Predictive Value that is introduced in Sec.~\ref{subsec: Appendix NPV}. As a consequence, a small value means the model is cautious is predicting observations negatively, while a large value means the opposite.

\subsubsection{Definition and Distribution}

The \textit{False Omission Rate} is commonly defined as
\begin{equation*}
  \FOR = \frac{\FN}{\TN + \FN}.
\end{equation*}
With the aid of Eq.~\eqref{eq: FN in TP}, this can be reformulated to
\begin{equation*}
  \FOR = \frac{P - \TP}{M - \MT}.
\end{equation*}
Note that this performance measure is only defined whenever $\MT < M$, otherwise the denominator is zero. Therefore, we assume specifically for $\FOR$ that $\theta < 1 - \frac{1}{2M}$. Now, $\FOR$ is linear in $\TP$ and can therefore be written as
\begin{align*}
  \FOR & = \Zab{\theta}{-\frac{1}{M-\MT}}{\frac{P}{M-\MT}} \sim \Hab{\theta}{-\frac{1}{M-\MT}}{\frac{P}{M-\MT}},
\end{align*}
with range:
\begin{align*}
  \FOR \stackrel{\eqref{eq: Range Zab}}{\in} \RZab{\theta}{-\frac{1}{M-\MT}}{\frac{P}{M-\MT}}.
\end{align*}

\subsubsection{Expectation}

Because $\FOR$ is linear in $\TP$ with slope $a = -1/(M-\MT)$ and intercept $b = P/(M-\MT)$, its expectation is
\begin{align*}
  \EE[\FOR] & = \EE\left[\Zab{\theta}{-\frac{1}{M-\MT}}{\frac{P}{M-\MT}}\right] \stackrel{\eqref{eq: Expectation Rule 1}}{=} -\frac{1}{M-\MT} \cdot \EE[\TP] + \frac{P}{M-\MT}
  = \frac{P}{M}.
\end{align*}

\subsubsection{Optimal Baselines}

The range of the expectation of $\FOR$ determines the baselines. The extreme values are defined as
\begin{align*}
  \min_{\theta \in [0, 1 - 1/(2M))} \left(\EE[\FOR]\right) & = \frac{P}{M}, \\
  \max_{\theta \in [0, 1 - 1/(2M))} \left(\EE[\FOR]\right) & = \frac{P}{M},
\end{align*}
because the expectation does not depend on $\theta$.
Consequently, the optimization values $\thmin$ and $\thmax$ are all allowed values for $\theta$:
\begin{equation*}
  \thmin = \thmax \in \left[0, 1 - \frac{1}{2M}\right).
\end{equation*}
This also means the discrete forms $\thsmin$ and $\thsmax$ of the optimizers are in the set of all allowed discrete values:
\begin{equation*}
  \thsmin = \thsmax \in \THSSPACE \setminus \{1\}.
\end{equation*}

\subsection{\texorpdfstring{$F_{\beta}$}{} Score} \label{subsec: Appendix FBETA}

The \textit{$F_{\beta}$ score} $\FB$ was introduced by \citet{Chinchor1992}. It is the weighted harmonic average between the True Positive Rate ($\TPR$) and the Positive Predictive Value ($\PPV$). These two performance measures are discussed extensively in Sec.~\ref{subsec: Appendix TPR} and~\ref{subsec: Appendix PPV}, respectively, and their summarized results are shown in Tables~\ref{tab: Properties baseline measures} and~\ref{tab: Optimal baseline measures}. The $F_{\beta}$ score balances predicting the actual positive observations correctly ($\TPR$) and being cautious in predicting observations as positive ($\PPV$). The factor $\beta > 0$ indicates how much more $\TPR$ is weighted compared to $\PPV$.

\subsubsection{Definition and Distribution}

The $F_{\beta}$ score is commonly defined as
\begin{equation*}
  \FB = \frac{1+\beta^2}{\frac{1}{\PPV} + \frac{\beta^2}{\TPR}}.
\end{equation*}
By using the definitions of $\TPR$ and $\PPV$ in Eq.~\eqref{eq:rawtprdef} and~\eqref{eq:rawppvdef}, $\FB$ can be formulated in terms of the base measures:
\begin{align*}
  \FB & = \frac{(1+\beta^2)\cdot\TP}{\beta^2 \cdot P + \TP + \FP}
\end{align*}
Eq.~\eqref{eq: TP in TP} and~\eqref{eq: FP in TP} allow us to write the formulation above in terms of only $\TP$:
\begin{align*}
  \FB = \frac{(1+\beta^2)\cdot\TP}{\beta^2 \cdot P + \MT}.
\end{align*}
Note that $P > 0$ and $\MT > 0$, otherwise $\TPR$ or $\PPV$ is not defined, and hence, $\FB$ is not defined. Now, $\FB$ is linear in $\TP$ and can be formulated as
\begin{align*}
  \FB & = \Zab{\theta}{\frac{1+\beta^2}{\beta^2\cdot P + \MT}}{0},
\end{align*}
with range:
\begin{align*}
  \FB \stackrel{\eqref{eq: Range Zab}}{\in} \RZab{\theta}{\frac{1+\beta^2}{\beta^2\cdot P + \MT}}{0}.
\end{align*}

\subsubsection{Expectation}

Because $\FB$ is linear in $\TP$ with slope $a = (1+\beta^2)/(\beta^2 P + \MT)$ and intercept $b = 0$, its expectation is given by
\begin{align}
  \EE[\FB] & = \EE\left[\Zab{\theta}{\frac{1+\beta^2}{\beta^2\cdot P + \MT}}{0}\right] \stackrel{\eqref{eq: Expectation Rule 1}}{=} \frac{1+\beta^2}{\beta^2\cdot P + \MT} \cdot \EE[\TP] + 0 = \frac{\MT \cdot P \cdot (1+\beta^2)}{M\cdot(\beta^2 \cdot P + \MT)} \nonumber \\
           & = \frac{(1+\beta^2)\cdot P\cdot \THS}{\beta^2\cdot P + M\cdot\THS}. \label{eq:appfbetae}
\end{align}

\subsubsection{Optimal Baselines}

To determine the extreme values of the expectation of $\FB$, and therefore the baselines, the derivative of the function $f: [0,1] \rightarrow [0,1]$ defined as
\begin{equation*}
  f(t) = \frac{(1+\beta^2) \cdot P \cdot t}{\beta^2 \cdot P + M \cdot t}
\end{equation*}
is calculated. First note that $\EE[\FB] = f(\MT / M)$. The derivative is given by
\begin{equation*}
  \frac{\mathrm{d}f(t)}{\mathrm{d}t} = \frac{\beta^2(1+\beta^2)\cdot P^2}{(\beta^2 \cdot P + M \cdot t)^2}.
\end{equation*}
It is strictly positive for all $t$ in its domain, thus $f$ is strictly increasing in $t$. This means $\EE[\FB]$ given in Eq.~\eqref{eq:appfbetae} is non-decreasing in both $\theta$ and $\THS$. This is because the term $\MT / M$ is non-decreasing in $\theta$.
Hence, the extreme values of the expectation of $\FB$ are its border values:
\begin{align*}
  \min_{\theta \in [1/(2M), 1]}\left(\EE[\FB]\right) & = \min_{\theta \in [1/(2M), 1]}\left(\frac{(1+\beta^2)\cdot P \cdot \MT}{M (\beta^2 \cdot P + \MT)} \right) = \frac{(1+\beta^2)\cdot P }{M(\beta^2\cdot P + 1)}, \\
  \max_{\theta \in [1/(2M), 1]}\left(\EE[\FB]\right) & = \max_{\theta \in [1/(2M), 1]}\left(\frac{(1+\beta^2)\cdot P \cdot \MT}{M (\beta^2 \cdot P + \MT)} \right) = \frac{(1+\beta^2) \cdot P}{\beta^2 \cdot P + M}.
\end{align*}
Consequently, the optimization values $\thmin$ and $\thmax$ for the extreme values are given by
\begin{align*}
  \thmin & \in \argmin_{\theta \in [1/(2M), 1]}\left(\EE[\FB]\right) = \argmin_{\theta \in [1/(2M), 1]}\left(\frac{\MT}{\beta^2 \cdot P + \MT}\right) = \begin{cases}[\frac{1}{2}, 1] & \text{if $M = 1$} \\ \left[\frac{1}{2M}, \frac{3}{2M}\right) & \text{if $M > 1$,} \end{cases}      \\
  \thmax & \in \argmax_{\theta \in [1/(2M), 1]}\left(\EE[\FB]\right) = \argmax_{\theta \in [1/(2M), 1]}\left(\frac{\MT}{\beta^2 \cdot P + \MT}\right) = \begin{cases}[\frac{1}{2}, 1] & \text{if $M = 1$} \\ \left[1 - \frac{1}{2M}, 1\right] & \text{if $M > 1$,} \end{cases}
\end{align*}
respectively. Following this reasoning, the discrete forms $\thsmin$ and $\thsmax$ are given by
\begin{align*}
  \thsmin & \in \argmin_{\THS \in \THSSPACE \setminus \{0\}}\left\{\EE[\FBS]\right\} = \argmin_{\THS \in \THSSPACE \setminus \{0\}}\left\{\frac{\THS}{\beta^2\cdot P + M\cdot\THS}\right\} = \left\{\frac{1}{M}\right\}, \\
  \thsmax & \in \argmax_{\THS \in \THSSPACE \setminus \{0\}}\left\{\EE[\FBS]\right\} = \argmax_{\THS \in \THSSPACE \setminus \{0\}}\left\{\frac{\THS}{\beta^2\cdot P + M\cdot\THS}\right\} = \{1\}.
\end{align*}

\subsection{Youden's J Statistic}\label{subsec: Appendix J}

The \textit{Youden's J Statistic} $\J$, \textit{Youden's Index}, or \textit{(Bookmaker) Informedness} was introduced by \citet{Youden1950} to capture the performance of a diagnostic test as a single statistic. It incorporates both the True Positive Rate and the True Negative Rate, which are discussed in Sec.~\ref{subsec: Appendix TPR} and~\ref{subsec: Appendix TNR}, respectively. Youden's J Statistic shows how well the model is able to correctly predict both the positive as the negative observations.

\subsubsection{Definition and Distribution}

The Youden's J Statistic is commonly defined as
\begin{equation*}
  \J = \TPR + \TNR - 1.
\end{equation*}
By using Eq.~\eqref{eq:tprdef} and~\eqref{eq:tnrdef}, which provide the definitions of $\TPR$ and $\TNR$ in terms of $\TP$, the definition of $\J$ can be reformulated as
\begin{align*}
  \J = \frac{M\cdot\TP - P\cdot\MT}{P\MPP}.
\end{align*}
Because $\TPR$ needs $P >0$, and $\TNR$ needs $N > 0$, we have both these assumptions for $\J$. Consequently, $M > 1$. Now, $\J$ is linear in $\TP$ and can therefore be written as
\begin{equation*}
  \J = \Zab{\theta}{\frac{M}{P\MPP}}{-\frac{\MT}{\MP}} \sim \Hab{\theta}{\frac{M}{P\MPP}}{-\frac{\MT}{\MP}},
\end{equation*}
with range:
\begin{align*}
  \J \stackrel{\eqref{eq: Range Zab}}{\in} \RZab{\theta}{\frac{M}{P\MPP}}{-\frac{\MT}{\MP}}.
\end{align*}

\subsubsection{Expectation}

Since $\J$ is linear in $\TP$ with slope $a = M/(P\MPP)$ and intercept $b = -\MT/\MPP$, its expectation is given by
\begin{align*}
  \EE[\J] & = \EE\left[\Zab{\theta}{\frac{M}{P\MPP}}{-\frac{\MT}{\MP}}\right] \stackrel{\eqref{eq: Expectation Rule 1}}{=} \frac{M}{P\MPP} \cdot \EE[\TP] -\frac{\MT}{\MP} = 0.
\end{align*}

\subsubsection{Optimal Baselines}

The extreme values of the expectation of $\J$ determine the baselines. They are given by
\begin{align*}
  \THMIN{\EE[\J]} & = 0, \\
  \THMAX{\EE[\J]} & = 0,
\end{align*}
because the expected value does not depend on $\theta$.
Consequently, the optimization values $\thmin$ and $\thmax$ can be any value in the domain of $\theta$:
\begin{equation*}
  \thmin = \thmax \in [0,1].
\end{equation*}
This also holds for the discrete forms $\thsmin$ and $\thsmax$ of the optimizers:
\begin{equation*}
  \thsmin = \thsmax \in \THSSPACE.
\end{equation*}

\subsection{Markedness} \label{subsec: Appendix MK}

The \textit{Markedness} $\MK$ or \textit{deltaP} is a performance measure that is mostly used in linguistics and social sciences. It combines both the Positive Predictive Value and the Negative Predictive Value. These two measures are discussed in Sec.~\ref{subsec: Appendix PPV} and~\ref{subsec: Appendix NPV}, respectively. The Markedness indicates how cautious the model is in predicting observations as positive and also how cautious it is in predicting them as negative.

\subsubsection{Definition and Distribution}

The Markedness is commonly defined as
\begin{equation*}
  \MK = \PPV + \NPV - 1.
\end{equation*}
This definition of $\MK$ can be reformulated in terms of $\TP$ by using Eq.~\eqref{eq:ppvdef} and~\eqref{eq:npvdef}:
\begin{equation*}
  \MK = \frac{M\cdot\TP - P\cdot\MT}{\MT(M-\MT)}.
\end{equation*}
Note that $\MK$ is only defined for $M>1$ and $\theta \in [1/(2M), 1 - 1/(2M))$, otherwise the denominator becomes zero. The assumption $M>1$ automatically follows from the assumptions $\hat{P} > 0$ and $\hat{N} > 0$, which hold for $\PPV$ and $\NPV$, respectively. In other words, there is at least one observation predicted positive and at least one predicted negative, thus $M > 1$. Now, $\MK$ is linear in $\TP$ and can therefore be written as
\begin{align*}
  \MK & = \Zab{\theta}{\frac{M}{\MT(M-\MT)}}{-\frac{P}{M-\MT}} \sim \Hab{\theta}{\frac{M}{\MT(M-\MT)}}{-\frac{P}{M-\MT}},
\end{align*}
with range:
\begin{align*}
  \MK & \stackrel{\eqref{eq: Range Zab}}{\in} \RZab{\theta}{\frac{M}{\MT(M-\MT)}}{-\frac{P}{M-\MT}}.
\end{align*}

\subsubsection{Expectation}

By using slope $a = M/(\MT(M-\MT))$ and intercept $b=-P/(M-\MT)$, the expectation of $\MK$ can be calculated:
\begin{align*}
  \EE[\MK] & = \EE\left[\Zab{\theta}{\frac{M}{\MT(M-\MT)}}{-\frac{P}{M-\MT}}\right] \stackrel{\eqref{eq: Expectation Rule 1}}{=} \frac{M}{\MT(M-\MT)} \cdot \EE[\TP] -\frac{P}{M-\MT} = 0.
\end{align*}

\subsubsection{Optimal Baselines}

The extreme values of the expectation of $\MK$ determine the baselines. Its range is given by:
\begin{align*}
  \min_{\theta \in [1/(2M), 1 - 1/(2M))} \left(\EE[\MK]\right) & = 0, \\
  \max_{\theta \in [1/(2M), 1 - 1/(2M))} \left(\EE[\MK]\right) & = 0,
\end{align*}
since the expected value does not depend on $\theta$. Therefore, the optimization values $\thmin$ and $\thmax$ are in the set of allowed values for $\theta$:
\begin{equation*}
  \thmin = \thmax \in \left[\frac{1}{2M}, 1 - \frac{1}{2M}\right).
\end{equation*}
This also means the discrete forms $\thsmin$ and $\thsmax$ of the optimizers are in the set of the allowed discrete values:
\begin{equation*}
  \thsmin = \thsmax \in \THSSPACE \setminus \{0,1\}.
\end{equation*}

\subsection{Accuracy} \label{subsec: Appendix ACC}

The \textit{Accuracy} $\ACC$ is the performance measure that assesses how good the model is in correctly predicting the observations without making a distinction between positive or negative observations.

\subsubsection{Definition and Distribution}

The Accuracy is commonly defined as
\begin{equation*}
  \ACC = \frac{\TP + \TN}{M}.
\end{equation*}
By using Eq.~\eqref{eq: TN in TP}, this can be restated as
\begin{align*}
  \ACC = \frac{2\cdot\TP + \MP - \MT}{M}.
\end{align*}
Note that it is linear in $\TP$ and can therefore be written as
\begin{equation}
  \ACC = \Zab{\theta}{\frac{2}{M}}{\frac{\MP-\MT}{M}} \sim \Hab{\theta}{\frac{2}{M}}{\frac{\MP-\MT}{M}}, \label{eq:acc}
\end{equation}
with range:
\begin{align*}
  \ACC \stackrel{\eqref{eq: Range Zab}}{\in} \RZab{\theta}{\frac{2}{M}}{\frac{\MP-\MT}{M}}.
\end{align*}

\subsubsection{Expectation}

Since $\ACC$ is linear in $\TP$ with slope $a = 2/M$ and intercept $b = (\MP-\MT)/M$, its expectation can be derived:
\begin{align}
  \EE[\ACC] & = \EE\left[\Zab{\theta}{\frac{2}{M}}{\frac{\MP-\MT}{M}}\right] \stackrel{\eqref{eq: Expectation Rule 1}}{=} \frac{2}{M} \cdot \EE[\TP] + \frac{\MP-\MT}{M} \nonumber \\
            & = \frac{(M-\MT)\MPP + \MT\cdot P}{M^2} = \frac{(1-\THS)\MPP + \THS\cdot P}{M}. \label{eq:expacc}
\end{align}

\subsubsection{Optimal Baselines}

The range of the expectation of $\ACC$ directly determines the baselines. To determine the extreme values of $\ACC$, the derivative of the function $f: [0,1] \rightarrow [0,1]$ defined as
\begin{equation*}
  f(t) = \frac{(1 - t)\MPP + P\cdot t}{M}
\end{equation*}
is calculated. First, note that $\EE[\ACC] = f(\MT / M)$. The derivative is given by
\begin{equation*}
  \frac{\mathrm{d}f(t)}{\mathrm{d}t} = \frac{2P - M}{M}.
\end{equation*}
It does not depend on $t$, but whether the derivative is positive or negative depends on $P$ and $M$. Whenever $P > \frac{M}{2}$, then $f$ is strictly increasing for all $t$ in its domain. If $P < \frac{M}{2}$, then $f$ is strictly decreasing. When $P = \frac{M}{2}$, $f$ is constant. Consequently, the same holds for $\EE[\ACC]$ given in Eq.~\eqref{eq:expacc}. This is because the term $\MT / M$ is non-decreasing in $\theta$. Thus, the extreme values of the expectation of $\ACC$ are given by
\begin{align*}
  \THMIN{\EE[\ACC]} & = \begin{cases} \frac{P}{M} & \text{if $P < \frac{M}{2}$} \\ 1 - \frac{P}{M} & \text{if $P \geq \frac{M}{2}$} \end{cases} = \min\left\{\frac{P}{M}, 1-\frac{P}{M}\right\}, \\
  \THMAX{\EE[\ACC]} & = \begin{cases} 1-\frac{P}{M} & \text{if $P < \frac{M}{2}$} \\ \frac{P}{M} & \text{if $P \geq \frac{M}{2}$} \end{cases} = \max\left\{\frac{P}{M}, 1-\frac{P}{M}\right\}.
\end{align*}
This means that the optimization values $\thmin \in [0,1]$ and $\thmax \in [0,1]$ for these extreme values respectively are given by
\begin{align}
  \thmin & \in \THARGMIN{\EE[\ACC]} = \begin{cases} \left[1 - \frac{1}{2M}, 1\right] & \text{if $P < \frac{M}{2}$} \\
              [0,1]                            & \text{if $P = \frac{M}{2}$} \\ \left[0, \frac{1}{2M}\right) & \text{if $P > \frac{M}{2}$},\end{cases} \label{eq:accmintheta}                                                                                \\
  \thmax & \in \THARGMAX{\EE[\ACC]} = \begin{cases} \left[0, \frac{1}{2M}\right) & \text{if $P < \frac{M}{2}$} \\ [0,1] & \text{if $P = \frac{M}{2}$} \\ \left[1 - \frac{1}{2M}, 1\right] & \text{if $P > \frac{M}{2}$}. \end{cases} \label{eq:accmaxtheta}
\end{align}
Consequently, the discrete versions $\thsmin \in \THSSPACE$ and $\thsmax \in \THSSPACE$ of the optimizers are given by
\begin{align}
  \thsmin & \in \THSARGMIN{\EE[\ACCS]} = \begin{cases} \{1\} & \text{if $P < \frac{M}{2}$} \\ \THSSPACE & \text{if $P = \frac{M}{2}$} \\ \{0\} & \text{if $P > \frac{M}{2}$,} \end{cases} \label{eq:accminths}  \\
  \thsmax & \in \THSARGMAX{\EE[\ACCS]} = \begin{cases} \{0\} & \text{if $P < \frac{M}{2}$} \\ \THSSPACE & \text{if $P = \frac{M}{2}$} \\ \{1\} & \text{if $P > \frac{M}{2}$,}  \end{cases} \label{eq:accmaxths}
\end{align}
respectively.

\subsection{Balanced Accuracy} \label{subsec: Appendix BACC}

The \textit{Balanced Accuracy} $\BACC$ is the mean of the True Positive Rate and True Negative Rate, which are discussed in Sec.~\ref{subsec: Appendix TPR} and~\ref{subsec: Appendix TNR}. It determines how good the model is in correctly predicting the positive observations and in correctly predicting the negative observations on average.

\subsubsection{Definition and Distribution}

The Balanced Accuracy is commonly defined as
\begin{equation*}
  \BACC = \frac{1}{2}\cdot(\TPR + \TNR).
\end{equation*}
By using Eq.~\eqref{eq:tprdef} and~\eqref{eq:tnrdef}, this can be reformulated as
\begin{align*}
  \BACC & = \frac{1}{2}\left(\frac{\TP}{P} + 1 - \frac{\MT - \TP}{\MP}\right) = \frac{M\cdot\TP}{2P \MPP} + \frac{\MP - \MT}{2\MPP}.
\end{align*}
Note that $P > 0$ and $N > 0$ should hold, otherwise $\TPR$ or $\TNR$ is not defined. Consequently, $M > 1$. Note that $\BACC$ is linear in $\TP$ and can therefore be written as
\begin{align*}
  \BACC & = \Zab{\theta}{\frac{M}{2P\MPP}}{\frac{\MP-\MT}{2\MPP}} \sim \Hab{\theta}{\frac{M}{2P\MPP}}{\frac{\MP-\MT}{2\MPP}},
\end{align*}
with range:
\begin{align*}
  \BACC \stackrel{\eqref{eq: Range Zab}}{\in} \RZab{\theta}{\frac{M}{2P\MPP}}{\frac{\MP-\MT}{2\MPP}}.
\end{align*}

\subsubsection{Expectation}

$\BACC$ is linear in $\TP$ with slope $a = M/(2P\MPP)$ and intercept $b = (\MP-\MT)/(2\MPP)$, so its expectation can be derived:
\begin{align*}
  \EE[\BACC] & = \EE\left[\Zab{\theta}{\frac{M}{2P\MPP}}{\frac{\MP-\MT}{2\MPP}}\right] \stackrel{\eqref{eq: Expectation Rule 1}}{=} \frac{M}{2P\MPP} \cdot \EE[\TP] + \frac{\MP-\MT}{2\MPP} = \frac{1}{2}.
\end{align*}

\subsubsection{Optimal Baselines}

The baselines are directly determined by the ranges of the expectation of $\BACC$. Since the expectation is constant, its extreme values are the same:
\begin{align*}
  \THMIN{\EE[\BACC]} & = \frac{1}{2}, \\
  \THMAX{\EE[\BACC]} & = \frac{1}{2}.
\end{align*}
This means that the optimization values $\thmin \in [0,1]$ and $\thmax \in [0,1]$ for these extreme values respectively are simply
\begin{align*}
  \thmin & \in \THARGMIN{\EE[\BACC]} = [0,1], \\
  \thmax & \in \THARGMAX{\EE[\BACC]} = [0,1].
\end{align*}
Consequently, the discrete versions $\thsmin \in \THSSPACE$ and $\thsmax \in \THSSPACE$ of the optimizers are given by
\begin{align*}
  \thsmin & \in \THSARGMIN{\EE[\ACCS]} = \THSSPACE, \\
  \thsmax & \in \THSARGMAX{\EE[\ACCS]} = \THSSPACE,
\end{align*}
respectively.

\subsection{Matthews Correlation Coefficient} \label{subsec: Appendix MCC}

The \textit{Matthews Correlation Coefficient} $\MCC$ was established by \citet{Matthews1975}. However, its definition is identical to that of the Yule phi coefficient, which was introduced by \citet{Yule1912}. The performance measure can be seen as the correlation coefficient between the actual and predicted classes. Hence, it is one of the few measures that lies in $[-1,1]$ instead of $[0,1]$.

\subsubsection{Definition and Distribution}

The Matthews Correlation Coefficient is commonly defined as
\begin{equation*}
  \MCC = \frac{\TP \cdot \TN - \FN \cdot \FP}{\sqrt{(\TP + \FP)(\TP + \FN)(\TN + \FP)(\TN + \FN)}}.
\end{equation*}
By using Eq.~\eqref{eq: FP in TP} and~\eqref{eq: TN in TP}, this definition can be reformulated as
\begin{equation}
  \MCC = \frac{M \cdot \TP - P \cdot \MT}{\sqrt{\MT \cdot P\MPP(M - \MT)}}. \label{eq:mccdef}
\end{equation}
As Table~\ref{tab: Assumptions P and N} shows, the assumptions $P > 0$, $N > 0$, $\hat{P} := \MT > 0$, and $\hat{N} := M - \MT > 0$ must hold. If one of these assumptions is violated, then the denominator in Eq.~\eqref{eq:mccdef} is zero, and $\MCC$ is not defined. Therefore, we have for $\MCC$ that $\frac{1}{2M} \leq \theta < 1 - \frac{1}{2M}$ and $M > 1$. Next, to improve readability we introduce the variable $C(M, P, \theta)$ to replace the denominator in Eq.~\eqref{eq:mccdef}:
\begin{align*}
  C(M, P, \theta) := \sqrt{\MT \cdot P\MPP(M - \MT)}.
\end{align*}
The definition of $\MCC$ is linear in $\TP$ and can thus be formulated as
\begin{align*}
  \MCC & = \Zab{\theta}{\frac{M}{C(M, P, \theta)}}{\frac{- P \cdot \MT}{C(M, P, \theta)}} \sim \Hab{\theta}{\frac{M}{C(M, P, \theta)}}{\frac{- P \cdot \MT}{C(M, P, \theta)}},
\end{align*}
with range:
\begin{align*}
  \MCC \stackrel{\eqref{eq: Range Zab}}{\in} \RZab{\theta}{\frac{M}{C(M, P, \theta)}}{\frac{- P \cdot \MT}{C(M, P, \theta)}}.
\end{align*}

\subsubsection{Expectation}

$\MCC$ is linear in $\TP$ with slope $a = M/C(M, P, \theta)$ and intercept $b = -P \cdot \MT/C(M, P, \theta)$, so its expectation can be derived from Eq.~\eqref{eq: Expectation Rule 1}:
\begin{align*}
  \EE[\MCC] & = \EE\left[\Zab{\theta}{\frac{M}{C(M, P, \theta)}}{\frac{- P \cdot \MT}{C(M, P, \theta)}}\right] \stackrel{\eqref{eq: Expectation Rule 1}}{=} \frac{M}{C(M, P, \theta)} \cdot \EE[\TP] - \frac{P \cdot \MT}{C(M, P, \theta)} =0.
\end{align*}

\subsubsection{Optimal Baselines}

The baselines are directly determined by the ranges of the expectation of $\MCC$. Since the expectation is constant, its extreme values are the same:
\begin{align*}
  \min_{\theta \in [1/(2M), 1 - 1/(2M))}\left(\EE[\MCC]\right) & = 0, \\
  \max_{\theta \in [1/(2M), 1 - 1/(2M))}\left(\EE[\MCC]\right) & = 0.
\end{align*}
This means that the optimization values $\thmin$ and $\thmax$ for these extreme values respectively are simply:
\begin{align*}
  \thmin = \thmax \in \left[\frac{1}{2M}, 1 - \frac{1}{2M}\right).
\end{align*}
Consequently, the discrete versions $\thsmin$ and $\thsmax$ of the optimizers are given by:
\begin{align*}
  \thsmin = \thsmax \in \THSSPACE \setminus \{0,1\}.
\end{align*}

\subsection{Cohen's Kappa} \label{subsec: Appendix KAPPA}

\textit{Cohen's kappa} $\KT$ is a less straightforward performance measure than the other measures that we discuss in this research. It is used to quantify the inter-rater reliability for two raters of categorical observations \citep{Kvlseth1989}. In our case, we compare the first rater, which is the \FancyName{} classifier, with the perfect rater, which assigns the true label to each observation.

\subsubsection{Definition and Distribution}

Although there are several definitions for Cohen's kappa, here we choose the following:
\begin{align*}
  \KT & = \dfrac{\PO - \PE}{1-\PE},
\end{align*}
with $\PO$ the Accuracy $\ACC$ as defined in Sec.~\ref{subsec: Appendix ACC} and $\PE$ the probability that the shuffle approach assigns the true label by chance. These two values can be expressed in terms of the base measures as follows:
\begin{align*}
  \PO & = \ACC = \frac{\TP +\TN}{M},                       \\
  \PE & = \frac{(\TP+\FP) \cdot P + (\TN + \FN)\MPP}{M^2}.
\end{align*}
By using Eq.~\eqref{eq:acc},~\eqref{eq: TP in TP},~\eqref{eq: FP in TP},~\eqref{eq: FN in TP} and~\eqref{eq: TN in TP} the above can be rewritten as
\begin{align*}
  \PO & = \frac{2\cdot\TP + \MP - \MT}{M},         \\
  \PE & = \frac{\MT \cdot P + (M - \MT)\MPP}{M^2}.
\end{align*}
Note that for $\KT$ to be well-defined, we need $1 - \PE \neq 0$. In other words,
\begin{align*}
  \MT \cdot P + (M - \MT)\MPP \neq M^2.
\end{align*}
This simplifies to
\begin{align}
  \frac{\MT}{M} \neq \frac{P}{2P-M}.
  \label{eq:kapparestriction}
\end{align}
The left-hand side is by definition in the interval $[0,1]$. For the right-hand side to be in that interval, we firstly need $P / (2P - M) \geq 0$. Since $P \geq 0$, that means $2P - M > 0$, and hence, $P > \frac{M}{2}$. Secondly, $P / (2P - M) \leq 1$. Since we know $P > \frac{M}{2}$, we obtain $P \geq M$. This inequality reduces to $P = M$, because $P$ is always at most $M$. Whenever $P = M$, then Eq.~\eqref{eq:kapparestriction} becomes
\begin{align*}
  \frac{\MT}{M} \neq 1.
\end{align*}
To summarize, when $P < M$, then all $\theta \in [0,1]$ are allowed in $\KT$, but when $P = M$, then $\theta < 1 - 1/(2M)$.

Now, by using $\PO$ and $\PE$ in the definition of Cohen's kappa, we obtain:
\begin{align*}
  \KT = \frac{2 \cdot M \cdot \TP - 2 \cdot \MT \cdot P}{P\left(M - \MT\right) + \MPP\MT}.
\end{align*}
To improve readability, we introduce the variables $a_{\KT}$ and $b_{\KT}$ defined as
\begin{align*}
  a_{\KT} & = \frac{2M}{P\left(M - \MT\right) + \MPP\MT}                   \\
  b_{\KT} & = -\frac{2\cdot \MT \cdot P}{P\left(M - \MT\right) + \MPP\MT}.
\end{align*}
Hence, $\KT$ is linear in $\TP$ and can be written as
\begin{equation*}
  \KT = \Zab{\theta}{a_{\KT}}{b_{\KT}} \sim \Hab{\theta}{a_{\KT}}{b_{\KT}},
\end{equation*}
with range:
\begin{align*}
  \KT \stackrel{\eqref{eq: Range Zab}}{\in} \RZab{\theta}{a_{\KT}}{b_{\KT}}.
\end{align*}

\subsubsection{Expectation}

As Cohen's kappa is linear in $\TP$, its expectation can be derived:
\begin{align*}
  \EE[\KT] & = \EE\left[\Zab{\theta}{a_{\KT}}{b_{\KT}}\right] \stackrel{\eqref{eq: Expectation Rule 1}}{=} a_{\KT} \cdot \EE[\TP] + b_{\KT} \\
           & = \frac{2\cdot \MT \cdot P}{P\left(M - \MT\right) + \MPP\MT} - \frac{2\cdot \MT \cdot P}{P\left(M - \MT\right) + \MPP\MT}      \\
           & = 0.
\end{align*}

\subsubsection{Optimal Baselines}

The baselines are directly determined by the ranges of the expectation of $\KT$. Since the expectation is constant, its extreme values are the same:
\begin{align*}
   & \begin{cases} \min_{\theta \in [0,1]}\left(\EE[\KT]\right) = 0        & \text{if $P < M$}  \\
              \min_{\theta \in [0,1-1/(2M))}\left(\EE[\KT]\right) = 0 & \text{if $P = M$,}\end{cases} \\
   & \begin{cases} \max_{\theta \in [0,1]}\left(\EE[\KT]\right) = 0        & \text{if $P < M$}  \\
              \max_{\theta \in [0,1-1/(2M))}\left(\EE[\KT]\right) = 0 & \text{if $P = M$.}\end{cases} \\
\end{align*}
This means that the optimization values $\thmin$ and $\thmax$ for these extreme values respectively are simply all allowed values:
\begin{align*}
  \begin{cases} \thmin = \thmax \in \left[0,1\right] & \text{if $P < M$} \\ \thmin = \thmax \in \left[0,1-\frac{1}{2M}\right] & \text{if $P = M$.} \end{cases}
\end{align*}
Consequently, the discrete versions $\thsmin$ and $\thsmax$ of the optimizers are given by
\begin{align*}
  \begin{cases} \thsmin = \thsmax \in \THSSPACE & \text{if $P < M$} \\ \thsmin = \thsmax \in \THSSPACE \setminus \{1\} & \text{if $P = M$.} \end{cases}
\end{align*}

\subsection{Fowlkes-Mallows Index} \label{subsec: Appendix FM}

The \textit{Fowlkes-Mallows Index} $\GONE$ or \textit{G-mean 1} was introduced by \citep{Fowlkes1983} as a way to calculate the similarity between two clusterings. It is the geometric average between the True Positive Rate ($\TPR$) and Positive Predictive Value ($\PPV$), which are discussed in Sec.~\ref{subsec: Appendix TPR} and~\ref{subsec: Appendix PPV}, respectively. It offers a balance between correctly predicting the actual positive observations ($\TPR$) and being cautious in predicting observations as positive ($\PPV$).

\subsubsection{Definition and Distribution}

The Fowlkes-Mallows Index is commonly defined as
\begin{align*}
  \GONE = \sqrt{\TPR \cdot \PPV}.
\end{align*}
By using the definitions of $\TPR$ and $\PPV$ in terms of $\TP$ in, respectively, Eq.~\eqref{eq:tprdef} and~\eqref{eq:ppvdef}, we obtain:
\begin{align*}
  \GONE = \frac{\TP}{\sqrt{P \cdot \MT}}.
\end{align*}
Since $\TPR$ is only defined when $P > 0$ and $\PPV$ only when $\hat{P} := \MT > 0$, also $\GONE$ has these assumptions. Therefore, $\theta \geq \frac{1}{2M}$.
The definition of $\GONE$ is linear in $\TP$ and can thus be formulated as
\begin{equation*}
  \GONE = \Zab{\theta}{\frac{1}{\sqrt{P \cdot \MT}}}{0} \sim \Hab{\theta}{\frac{1}{\sqrt{P \cdot \MT}}}{0},
\end{equation*}
with range:
\begin{align*}
  \GONE \stackrel{\eqref{eq: Range Zab}}{\in} \RZab{\theta}{\frac{1}{\sqrt{P \cdot \MT}}}{0}.
\end{align*}

\subsubsection{Expectation}

Because $\GONE$ is linear in $\TP$ with slope $a = 1/\sqrt{P\cdot\MT}$ and intercept $b = 0$, its expectation is
\begin{align*}
  \EE[\GONE] & = \EE\left[\Zab{\theta}{\frac{1}{\sqrt{P \cdot \MT}}}{0}\right] \stackrel{\eqref{eq: Expectation Rule 1}}{=} \frac{1}{\sqrt{P \cdot \MT}} \cdot \EE[\TP] + 0 = \frac{\sqrt{P \cdot \MT}}{M} = \sqrt{\frac{\THS \cdot P}{M}}.
\end{align*}

\subsubsection{Optimal Baselines}

The extreme values of the expectation of $\GONE$ determine the baselines. They are given by:
\begin{align*}
  \min_{\theta \in [1/(2M), 1]} \left(\EE[\GONE]\right) & = \min_{\theta \in [1/(2M), 1]} \left( \frac{\sqrt{P \cdot \MT}}{M}\right) = \frac{\sqrt{P}}{M}, \\
  \max_{\theta \in [1/(2M), 1]} \left(\EE[\GONE]\right) & = \max_{\theta \in [1/(2M), 1]} \left( \frac{\sqrt{P \cdot \MT}}{M}\right) = \sqrt{\frac{P}{M}},
\end{align*}
because the expectation is a non-decreasing function in $\theta$. Note that the minimum and maximum are equal to each other when $M = 1$.
Consequently, the optimizers $\thmin$ and $\thmax $ for the extreme values are determined by:
\begin{align*}
  \thmin & \in \argmin_{\theta \in [1/(2M), 1]} \left(\EE[\GONE]\right) = \argmin_{\theta \in [1/(2M), 1]}\left(\frac{\sqrt{P \cdot \MT}}{M}\right) = \begin{cases} \left[\frac{1}{2M}, 1\right] & \text{if $M = 1$} \\ \left[\frac{1}{2M}, \frac{3}{2M}\right) & \text{if $M > 1$}, \end{cases}
\end{align*}
\begin{align*}
  \thmax & \in \argmax_{\theta \in [1/(2M), 1]} \left(\EE[\GONE]\right) = \argmax_{\theta \in [1/(2M), 1]}\left(\frac{\sqrt{P \cdot \MT}}{M}\right) = \begin{cases} \left[\frac{1}{2M}, 1\right] & \text{if $M = 1$} \\ \left[1 - \frac{1}{2M}, 1\right] & \text{if $M > 1$}, \end{cases}
\end{align*}
respectively. The discrete forms $\thsmin$ and $\thsmax$ of these are given by:
\begin{align*}
  \thsmin & \in \argmin_{\THS \in \THSSPACE \setminus \{0\}} \left\{\EE[\GONES]\right\} = \argmin_{\THS \in \THSSPACE \setminus \{0\}} \left\{\sqrt{\frac{\THS \cdot P}{M}}\right\} = \left\{\frac{1}{M}\right\}, \\
  \thsmax & \in \argmax_{\THS \in \THSSPACE \setminus \{0\}} \left\{\EE[\GONES]\right\} = \argmax_{\THS \in \THSSPACE \setminus \{0\}} \left\{\sqrt{\frac{\THS \cdot P}{M}}\right\} = \{1\}.
\end{align*}

\subsection{G-mean 2}\label{subsec: Appendix G2}

The \textit{G-mean 2} $\GTWO$ was established by \citep{Kubat1998}. This performance measure is the geometric average between the True Positive Rate ($\TPR$) and True Negative Rate ($\TNR$), which we discuss in Sec.~\ref{subsec: Appendix TPR} and~\ref{subsec: Appendix TNR}, respectively. Hence, it balances correctly predicting the positive observations and correctly predicting the negative observations.

\subsubsection{Definition and Distribution}

The G-mean 2 is defined as
\begin{align*}
  \GTWO = \sqrt{\TPR \cdot \TNR}.
\end{align*}
Since $\TPR$ needs the assumption $P > 0$ and $\TNR$ needs $N := M - P > 0$, we have these restrictions also for $\GTWO$. Consequently, $M > 1$. Now, by using the definitions of $\TPR$ and $\TNR$ in terms of $\TP$ in, respectively, Eq.~\eqref{eq:tprdef} and~\eqref{eq:tnrdef}, we obtain:
\begin{align*}
  \GTWO & = \sqrt{\frac{\TP \cdot (\MP - \MT) + \TP^2}{P\MPP}}.
\end{align*}
This function is not a linear function of $\TP$, and hence, we cannot write it in the form $\Zab{\theta}{a}{b} = a \cdot \TP + b$ for some variables $a, b \in \mathbb{R}$.

\subsubsection{Expectation}

Since $\GTWO$ is not linear in $\TP$, we cannot easily use the expectation of $\TP$ to determine that for $\GTWO$.
However, we are able to determine the second moment of $\GTWO$:
\begin{align*}
  \EE\left[\left(\GTWO\right)^2\right] & = \frac{M - P - \MT}{P \MPP} \cdot \EE[\TP] + \frac{1}{P \MPP} \cdot \EE[\TP^2] \nonumber                                          \\
                                       & = \frac{M - P - \MT}{P \MPP} \cdot \frac{\MT}{M} \cdot P + \frac{1}{P \MPP} \cdot \left(\Var[\TP] + \EE[\TP]^2\right)) \nonumber   \\
                                       & = \frac{(M-P-\MT)\cdot\MT}{M\MPP} + \frac{\frac{\MT(M-\MT)P\MPP}{M^2(M-1)} + \left(\frac{\MT}{M}\cdot P\right)^2}{P\MPP} \nonumber \\
                                       & = \frac{\MT\cdot(M-\MT)}{M(M-1)}
  = \THS \cdot \left(1 - \THS\right) \cdot \frac{M}{M-1}. %
\end{align*}
Of course, since the distribution of $\TP$ is known, the expectation of $\GTWO$ can always be numerically calculated.

\subsubsection{Optimal Baselines}

Since the function $\varphi: \mathbb{R} \rightarrow \mathbb{R}_{\geq 0}$ given by $\varphi(x) = x^2$ is a convex function, we have by Jensen's inequality that
\begin{align*}
  \EE[\GTWO]^2 \leq \EE\left[\left(\GTWO\right)^2\right] = \THS \left(1 - \THS\right) \frac{M}{M-1}.
\end{align*}
This means that
\begin{equation*}
  \EE[\GTWO] \leq \sqrt{\THS\left(1 - \THS\right) \frac{M}{M-1}}.
\end{equation*}
Therefore, whenever $\THS \in \{0,1\}$, then $\EE[\GTWO] \leq 0$. Since $\GTWO \geq 0$, it must hold that $\EE[\GTWO] = 0$. Hence, the set $\{0,1\}$ contains minimizers for $\EE[\GTWO]$. The continuous version of this set is the interval $[0, 1/(2M)) \cup [1 - 1/(2M), 1]$. To show that this interval contains the only possible values for the minimizers, consider the definition for the expectation of $\GTWO$:
\begin{align*}
  \EE\left[\GTWO\right] & = \sum_{k \in \mathcal{D}(\TP)} \sqrt{\frac{k\cdot\left(\MPP - (\MT - k)\right)}{P\MPP}}\cdot\PP(\TP = k),
\end{align*}
where $\mathcal{D}(\TP)$ is the domain of $\TP$, i.e., the set of values $k$ such that $\PP(\TP = k) > 0$. Now, let $\theta$ be such that $1/(2M) \leq \theta < 1 - 1/(2M)$. Furthermore, consider the summand  $S^{(\theta)}_k$ corresponding to $k = \min\{P, \MT\} \in \mathcal{D}(\TP)$:
\begin{align*}
  S^{(\theta)}_{k = \min\{P, \MT\}} & = \begin{cases}  \sqrt{\frac{M-\MT}{M-P}} \cdot\PP(\TP = P)  & \text{if $P \leq \MT$} \\ \sqrt{\frac{\MT}{P}} \cdot\PP(\TP = \MT) & \text{ if $P > \MT$,} \end{cases}
\end{align*}
which is strictly positive in both cases. Hence, there is at least one term in the summation in the definition of $\EE\left[\GTWO\right]$ that is larger than 0, thus the expectation is strictly positive for $1/(2M) \leq \theta < 1 - 1/(2M)$. Consequently, the minimization values $\thmin \in [0,1]$ are
\begin{align*}
  \thmin & \in \THARGMIN{\EE[\GTWO]} = \left[0, \frac{1}{2M}\right) \cup \left[1 - \frac{1}{2M}, 1\right].
\end{align*}
Following this reasoning, the discrete form $\thsmin \in \THSSPACE$ is given by
\begin{align*}
  \thsmin & \in \THSARGMIN{\EE[\GTWO]} = \{0, 1\}.
\end{align*}

\subsection{Prevalence Threshold (PT)}\label{subsec: Appendix PT}

A relatively new performance measure named \emph{Prevalence Threshold} ($\PT$) was introduced by \citep{Balayla2020}. We could not find many articles that use this measure, but it is included for completeness. However, this performance measure has an inherent problem that eliminates the possibility to determine all statistics.

\subsubsection{Definition and Distribution}

The \textit{Prevalence Threshold} $\PT$ is commonly defined as
\begin{align*}
  \PT & = \frac{\sqrt{\TPR \cdot \FPR} - \FPR}{\TPR - \FPR}.
\end{align*}
By using the definitions of $\TPR$ and $\FPR$ in terms of $\TP$ (see Equations~\eqref{eq:tprdef} and~\eqref{eq:fprdef}), we obtain:
\begin{align}
  \PT & = \frac{\sqrt{P\cdot\MPP\cdot \TP \cdot (\MT - \TP)}-P(\MT - \TP)}{M\cdot\TP - P \cdot \MT}.
  \label{eq: Distribution PT}
\end{align}
It is clear that this performance measure is not a linear function of $\TP$, therefore we cannot easily calculate its expectation. However, there are more fundamental problems with $\PT$.

\subsubsection{Division by Zero}
Eq.~\eqref{eq: Distribution PT} shows that $\PT$ is a problematic measure. When is the denominator zero? This happens when $\TP = (\MT / M) \cdot P$. In this case, the fraction is undefined, as the denominator is zero. Furthermore, also the numerator is zero in that case. The number of true positives $\TP$ can attain the value $(\MT / M) \cdot P = \THS \cdot P$ whenever the latter is also an integer. For example, this always happens for $\THS \in \{0, 1\}$. But even when $\THS \in \THSSPACE \setminus \{0, 1\}$, $\PT$ is still only safe to use when $M$ and $P$ are coprime, i.e., when the only positive integer that is a divisor of both of them is 1. Otherwise, there are always values of $\THS \in \THSSPACE \setminus \{0, 1\}$ that cause $\THS \cdot P$ to be an integer and therefore $\PT$ to be undefined when $\TP$ attains that value.

One solution would be to say $\PT := c$, $c \in [0,1]$, whenever both the numerator and denominator are zero. However, this $c$ is arbitrary and directly influences the optimization of the expectation. This makes the optimal parameter values dependent on $c$, which is beyond the scope of this chapter. Thus, no statistics are derived for the Prevalence Threshold $\PT$.

\subsection{Threat Score (TS) / Critical Success Index (CSI)}\label{subsec: Appendix TS}

The \textit{Threat Score} \citep{Palmer1949} $\TS$ or \textit{Critical Success Index} \citep{schaefer1990} is a performance measure that is used for evaluation of forecasting binary weather events: it either happens in a specific location or it does not. It was already used in 1884 to evaluate the prediction of tornadoes~\citep{schaefer1990}. The Threat Score is the ratio of successful event forecasts ($\TP$) to the total number of positive predictions ($\TP + \FP$) and the number of events that were missed ($\FN$).

\subsubsection{Definition and Distribution}

The Threat Score is thus defined as
\begin{align*}
  \TS & = \frac{\TP}{\TP + \FP + \FN}.
\end{align*}
By using Eq.~\eqref{eq: FP in TP} and~\eqref{eq: FN in TP}, this definition can be reformulated as
\begin{align*}
  \TS & = \frac{\TP}{P + \MT - \TP}.
\end{align*}
Note that $\TS$ is well-defined whenever $P > 0$. The definition of $\TS$ is not linear in $\TP$, and so there are no $a, b \in \mathbb{R}$ such that we can write the definition as $\Zab{\theta}{a}{b}$. %

\subsubsection{Expectation}

Because $\TS$ is not linear in $\TP$, determining the expectation is less straightforward than for other performance measures. The definition of the expectation is
\begin{align*}
  \EE[\TS] & = \sum_{k\in \mathcal{D}(\TP)}\frac{k}{P + \MT - k} \cdot \PP(\TP = k).
\end{align*}
Unfortunately, we cannot explicitly solve this sum, but it can be calculated numerically.

\subsubsection{Optimal Baselines}

Although no explicit formula can be given for the expectation, we are able to calculate the extreme values of the expectation and the corresponding optimizers.

\paragraph{Minimal Baseline}
Firstly, we show that $\thmin \in [0, \frac{1}{2M})$ constitutes a minimum and that there are no $\theta$ outside this interval also yielding this minimum. To this end,
\begin{align*}
  \EE[\text{TS}_{\thmin}] & = \sum_{k\in \mathcal{D}(\text{TS}_{\thmin})}\frac{k}{P + 0 - k} \cdot \PP(\text{TS}_{\thmin} = k) = 0,
\end{align*}
because $\mathcal{D}(\text{TS}_{\thmin})= \{0\}$. This is the lowest possible value, since $\TS$ is a non-negative performance measure, and hence, $\EE[\TS] \geq 0$ for any $\theta \in [0,1]$. Now, let $\theta' \geq \frac{1}{2M}$, then there exists a $k' > 0$ such that $\PP(\text{TP}_{\theta'} = k') > 0$. Consequently, $\EE[\text{TS}_{\theta'}] > 0$ and this means the interval $[0, \frac{1}{2M})$ contains the only values that constitute the minimum. In summary,
\begin{align*}
  \THMIN{\EE[\TS]}               & = 0,                            \\
  \thmin \in \THARGMIN{\EE[\TS]} & = \left[0, \frac{1}{2M}\right).
\end{align*}
Since $\thsmin$ is the discretization of $\thmin$ it corresponds to 0. More precisely:
\begin{align*}
  \thsmin \in \THSARGMIN{\EE[\TSS]} = \{0\}.
\end{align*}

\paragraph{Maximal Baseline}
Secondly, to determine the maximum of $\EE[\TS]$ and the corresponding parameter $\thmax$, we determine an upper bound for the expectation, show that this value is attained for a specific interval and that there is no $\theta$ outside this interval also yielding this value. To do this, assume that $\MT > 0$. This makes sense, because $\MT = 0$ implies $\theta < 1/(2M)$ and such a $\theta$ would yield the minimum $0$. Now,
\begin{align*}
  \EE[\TS] & = \sum_{k\in \mathcal{D}(\TP)}\frac{k}{P + \MT - k} \cdot \PP(\TP = k)                                                                                                                                                      \\
           & \leq \sum_{k\in \mathcal{D}(\TP)}\frac{k}{P + \MT - P} \cdot \PP(\TP = k) = \frac{1}{\MT} \sum_{k\in \mathcal{D}(\TP)}k \cdot \PP(\TP = k) = \frac{\EE[\TP]}{\MT} \stackrel{\eqref{eq: Expectation Rule 1}}{=} \frac{P}{M}.
\end{align*}
Next, let $\thmax \in [1 - 1/(2M), 1]$, then
\begin{align*}
  \EE[\text{TS}_{\thmax}] & = \sum_{k=M - (M-P)}^{P}\frac{k}{P + M - k} \cdot \PP(\text{TP}_{\thmax} = k) = \frac{P}{P + M - P} \cdot \PP(\text{TP}_{\thmax} = P) = \frac{P}{M},
\end{align*}
because $\PP(\text{TP}_{\thmax} = P) = 1$. Hence, the upper bound is attained for $\thmax \in [1 - 1/(2M), 1]$, and thus, $\thmax$ is a maximizer.

Now, specifically for $P = 1$, we show that the interval of maximizers is actually $[1/(2M), 1]$. Thus, let $\theta \in [1/(2M), 1-1/(2M))$, then $0 < \MT < M$ and
\begin{align*}
  \EE[\TS] & = \sum_{k=\max\{0, \MT - (M-1)\}}^{\min\{1, \MT\}}\frac{k}{1 + \MT - k} \cdot \PP(\TP = k)                             \\
           & = \frac{0}{1 + \MT - 0} \cdot \PP(\TP = 0) + \frac{1}{1 + \MT - 1} \cdot \PP(\TP = 1) = \frac{1}{\MT}\cdot\PP(\TP = 1) \\
           & = \frac{1}{\MT} \cdot \left(\frac{\binom{1}{1} \binom{M-1}{\MT-1}}{\binom{M}{\MT}}\right) = \frac{1}{M},
\end{align*}
which is exactly the upper bound $\EE[\text{TS}_{\thmax}] = P/M$ for $P = 1$.

Next, to show that the maximizers are only in $[1 - 1/(2M), 1]$ for $P > 1$, assume there is a $\theta' < 1-\frac{1}{2M}$ that also yields the maximum. Hence, there is a $k' \in \mathcal{D}(\text{TP}_{\theta'})$ with $0 < k' < P$ such that $\PP(\text{TP}_{\theta'} = k')$. This means
\begin{align*}
  \EE[\text{TS}_{\theta'}] & = \sum_{k \in \mathcal{D}(\text{TP}_{\theta'})} \frac{k}{P + \lfloor M \cdot \theta' \rceil - k} \cdot \PP(\text{TP}_{\theta'} = k)                                                                                                               \\
                           & = \frac{k'}{P + \lfloor M \cdot \theta' \rceil - k'} \cdot \PP(\text{TP}_{\theta'} = k') + \sum_{k \in \mathcal{D}(\text{TP}_{\theta'})\setminus\{k'\}} \frac{k}{P + \lfloor M \cdot \theta' \rceil - k} \cdot \PP(\text{TP}_{\theta'} = k)       \\
                           & \leq \frac{k'}{P + \lfloor M \cdot \theta' \rceil - (P-1)} \cdot \PP(\text{TP}_{\theta'} = k') + \sum_{k \in \mathcal{D}(\text{TP}_{\theta'})\setminus\{k'\}} \frac{k}{P + \lfloor M \cdot \theta' \rceil - P} \cdot \PP(\text{TP}_{\theta'} = k) \\
                           & \quad = \frac{k'}{\lfloor M \cdot \theta' \rceil + 1} \PP(\text{TP}_{\theta'} = k') + \sum_{k \in \mathcal{D}(\text{TP}_{\theta'})\setminus\{k'\}} \frac{k}{\lfloor M \cdot \theta' \rceil} \PP(\text{TP}_{\theta'} = k)                          \\
                           & \quad < \frac{k'}{\lfloor M \cdot \theta' \rceil} \cdot \PP(\text{TP}_{\theta'} = k') + \sum_{k \in \mathcal{D}(\text{TP}_{\theta'})\setminus\{k'\}} \frac{k}{\lfloor M \cdot \theta' \rceil} \cdot \PP(\text{TP}_{\theta'} = k)                  \\
                           & \qquad = \frac{1}{\lfloor M \cdot \theta' \rceil}\sum_{k \in \mathcal{D}(\text{TP}_{\theta'})} k \cdot \PP(\text{TP}_{\theta'} = k) = \frac{P}{M}.
\end{align*}
Hence, there is a strict inequality $\EE[\text{TS}_{\theta'}] < \frac{P}{M}$ and this means $\theta'$ is not a maximizer of the expectation. Consequently, the maximizers are only in the interval $[1 - 1/(2M), 1]$ for $P > 1$. In summary,
\begin{align*}
  \THMAX{\EE[\TS]}               & = \frac{P}{M},                                                                                                                                           \\
  \thmax \in \THARGMAX{\EE[\TS]} & = \begin{cases}\left[\frac{1}{2M}, 1\right] & \text{if $P=1$} \\ \left[1-\frac{1}{2M}, 1\right] & \text{if $P > 1$}. \end{cases}
\end{align*}
Since $\thsmax$ is the discretization of $\thmax$, we obtain:
\begin{align*}
  \thsmax \in \THSARGMAX{\EE[\TSS]} = \begin{cases}\THSSPACE \setminus \{0\} & \text{if $P=1$} \\ \{1\} & \text{if $P > 1$}. \end{cases}
\end{align*}

\paragraph{Declarations}
\begin{itemize}
  \item \textbf{Funding:} No funding was received for conducting this study.
  \item \textbf{Conflicts of interest/Competing interests:} Not applicable.
  \item \textbf{Availability of data and material:} All data used in this research is cited in the appropriate sections.
  \item \textbf{Code availability:} The Dutch Draw code can be found at \url{https://github.com/joris-pries/DutchDraw}.
  \item \textbf{Authors' contributions} (\textit{Contributor Roles Taxonomy} (CRediT)):
        \begin{itemize}
          \item Etienne van de Bijl: Conceptualization, Methodology, Software, Validation, Formal analysis, Investigation, Data Curation, Writing - Original Draft, Writing - Review \& Editing, Visualization, Project administration;
          \item Jan Klein: Conceptualization, Methodology, Software, Validation, Formal analysis, Investigation, Data Curation, Writing - Original Draft, Writing - Review \& Editing, Visualization, Project administration;
          \item Joris Pries: Conceptualization, Methodology, Software, Validation, Formal analysis, Investigation, Data Curation, Writing - Original Draft, Writing - Review \& Editing, Visualization, Project administration;
          \item Sandjai Bhulai: Conceptualization, Validation, Writing - Review \& Editing, Supervision;
          \item Mark Hoogendoorn: Conceptualization, Validation, Writing - Review \& Editing, Supervision;
          \item Rob van der Mei: Conceptualization, Validation, Writing - Review \& Editing, Supervision.
        \end{itemize}
\end{itemize}

\bibliographystyle{spbasic-edit}      %

\end{document}